\documentclass[twoside,11pt]{article}
\usepackage[dvipdfmx]{graphicx}
\usepackage{blindtext, epsdice}

\usepackage[preprint]{jmlr2e}

\usepackage{amsmath}
\usepackage[normalem]{ulem}
\usepackage{cleveref}
\usepackage{xcolor}
\usepackage{wrapfig}
\usepackage{enumitem}
\newcommand{\vparam}{\boldsymbol{\theta}}

\newcommand\cut[1]{}

\newcommand{\squishlist}{
   \begin{list}{$\bullet$}
    { \setlength{\itemsep}{0pt}      \setlength{\parsep}{3pt}
      \setlength{\topsep}{3pt}       \setlength{\partopsep}{0pt}
      \setlength{\leftmargin}{1.5em} \setlength{\labelwidth}{1em}
      \setlength{\labelsep}{0.5em} } }

\newcommand{\squishlisttwo}{
   \begin{list}{$\bullet$}
    { \setlength{\itemsep}{0pt}    \setlength{\parsep}{0pt}
      \setlength{\topsep}{0pt}     \setlength{\partopsep}{0pt}
      \setlength{\leftmargin}{2em} \setlength{\labelwidth}{1.5em}
      \setlength{\labelsep}{0.5em} } }

\newcommand{\squishend}{
    \end{list}  }

{}
{}
{}
{}
\newcommand{\sign}{\mbox{sign}}

\newcommand{\myvec}[1]{\mbox{$\mathbf{#1}$}}
\newcommand{\myvecsym}[1]{\mbox{$\boldsymbol{#1}$}}

\newcommand{\vlambda}{\mbox{$\myvecsym{\lambda}$}}

\newcommand{\vv}{\mbox{$\myvec{v}$}}
\newcommand{\vw}{\mbox{$\myvec{w}$}}

\newcommand{\vx}{\mbox{$\myvec{x}$}}

\newcommand{\vy}{\mbox{$\myvec{y}$}}

\newcommand{\vF}{\mbox{$\myvec{F}$}}

\newcommand{\vS}{\mbox{$\myvec{S}$}}
\newcommand{\vT}{\mbox{$\myvec{T}$}}

\newcommand{\be}{\begin{equation}}
\newcommand{\ee}{\end{equation}}
\newcommand{\bea}{\begin{eqnarray}}
\newcommand{\eea}{\end{eqnarray}}
\newcommand{\beaa}{\begin{eqnarray*}}
\newcommand{\eeaa}{\end{eqnarray*}}

\newcommand{\bbR}{\mathbb{R}}

\DeclareMathOperator*{\argmin}{arg\,min}

\newcommand{\dualsp}{\mathrm{V}^*}
\newcommand{\dual}{^{*}}
\newcommand{\primsp}{\mathrm{V}}
\newcommand{\dd}{{\mathrm{d}}}
\newcommand{\soln}[1]{{\hat{#1}}}

\renewcommand{\vec}[1]{{\mathbf{\boldsymbol{#1}}}}

\newcommand{\llambda}{{\boldsymbol{\lambda}}}
\newcommand{\mmu}{{\boldsymbol{\mu}}}

\newcommand{\ttheta}{{\boldsymbol{\theta}}}
\newcommand{\aalpha}{{\boldsymbol{\alpha}}}
\newcommand{\eeta}{{\boldsymbol{\eta}}}
\newcommand{\zzeta}{{\boldsymbol{\zeta}}}

\newcommand{\DKL}{\operatorname{KL}}
\newcommand{\R}{\mathbb{R}}
\newcommand{\EE}{\mathbb{E}}
\DeclareMathOperator{\EF}{EF}
\DeclareMathOperator{\Stab}{Stab}
\DeclareMathOperator{\Aff}{Aff}
\DeclareMathOperator{\GL}{GL}
\DeclareMathOperator{\SL}{SL}
\DeclareMathOperator{\SO}{SO}
\newcommand{\Qcal}{\mathcal{Q}}
\newcommand{\inv}{^{-1}}
\newcommand{\Sym}{\operatorname{Sym}}
\newcommand{\pairing}[2]{\left\langle #1, #2 \right\rangle}
\DeclareMathOperator*{\argmax}{\arg\,\max}
\DeclareMathOperator{\Proj}{Proj}
\DeclareMathOperator{\Ad}{Ad}
\DeclareMathOperator{\Tr}{Tr}

\newtheorem{setup}[theorem]{Setup}

\definecolor{TolMutedBlue}{HTML}{332288}
\definecolor{TolMutedGreen}{HTML}{117733}
\definecolor{TolMutedPurple}{HTML}{AA4499}

\hypersetup{
    colorlinks=true,
    citecolor=TolMutedBlue,%
    linkcolor=TolMutedGreen,%
    urlcolor=TolMutedPurple,% 
}

\crefname{section}{Sec.}{Sec.}
\crefname{thm}{Thm.}{Theorem}
\crefname{appendix}{App.}{Appendices}
\crefname{algorithm}{Alg.}{Algorithms}
\crefname{equation}{Eq.}{Eqs.}
\crefname{figure}{Fig.}{Figs.}
\crefname{prop}{Prop.}{Props.}
\creflabelformat{equation}{#2\textup{#1}#3} % Remove the brackets around Equation numbers

\usepackage{lastpage}
\ShortHeadings{Variational Learning with Lie Group Exponential Families}{E. M. K{\i}ral,  T. M\"ollenhoff,  K. Tojo, M. E. Khan, K. Bannai}
\firstpageno{1}

\begin{document}

%\title{Lie Group Structure of Exponential Families for Variational Learning}
\title{Variational Learning with Lie Group Exponential Families}

\author{\name Eren Mehmet K{\i}ral \email erenmehmetkiral@protonmail.com \\
\addr RIKEN Center for Advanced Intelligence Project\\
	   1-4-1 Nihonbashi, Chuo-ku, Tokyo 103-0027, Japan
       \AND
       \name Thomas M\"ollenhoff \email thomas.moellenhoff@riken.jp \\
       \addr RIKEN Center for Advanced Intelligence Project\\
	   1-4-1 Nihonbashi, Chuo-ku, Tokyo 103-0027, Japan
\AND
\name Koichi Tojo \email koichi.tojo@tokai.ac.jp \\
\addr Department of Mathematical Sciences, Tokai University, \\
	   4-1-1 Kitakaname, Hiratsuka-shi, Kanagawa 259-1292, Japan
       \AND
       \name Mohammad Emtiyaz Khan \email emtiyaz.khan@riken.jp \\
       \addr Department of Computer Science, Technische Universität Darmstadt\\
       RIKEN Center for Advanced Intelligence Project\\
	   Hochschulstraße 10, Darmstadt, 64289, Hessen, Germany 
       \AND
       \name Kenichi Bannai \email bannai@math.keio.ac.jp \\
       \addr Department of Mathematics, Keio University, \\
	   RIKEN Center for Advanced Intelligence Project \\
	   3-14-1 Hiyoshi, Kohoku-ku, Yokohama, 223-8522, Japan
}

\editor{(to be decided)}

\maketitle

\begin{abstract}%   <- trailing '%' for backward compatibility of .sty file
%	Variational Bayesian learning for deep networks offers many potential benefits 
%	but formulating good algorithms is not straightforward. This is due to constraints on the variational distribution's parameters which can be difficult 
%	to enforce and a need for low-variance gradient estimators which are not always easily available. 
%	Recently proposed Lie group methods address the above difficulties but their relation 
%	to widely used natural-gradient variational learning methods such as the Bayesian learning rule (BLR) is not fully understood.
%	In this paper we show that linear approximations 
%	to the Lie group algorithm generalize the BLR. 
%	The BLR is derived for exponential families, yet such a connection is possible 
%	due to recent advances that show how most common  
%	exponential families have an underlying Lie group structure.
%	Our work provides a way to construct
%	efficiently implementable algorithms for variational learning with common
%	exponential families. 
Variational Bayesian learning for deep neural networks offers many potential benefits.
The difficulty in implementation arises from constraints on the parameters of the variational distributions, as well as the need for low-variance gradient estimators. Our recent work proposes Lie group methods for multiplicative or affine groups that naturally overcome such difficulties. Lie groups are essentially parameter spaces with a binary operator such as addition or multiplication, which allows for natural updates within the parameter space. In this article, we streamline and extend our Lie group framework to any general Lie group, and give an explicit relation to standard natural-gradient variational learning methods such as the Bayesian learning rule (BLR). In particular, we prove that linear approximations of the Lie group algorithm give generalizations of the BLR. The key ingredient is the harmonic exponential family, a general class of Lie group exponential families associated to representations of Lie groups. Our work provides a systematic method to construct from such representations efficiently implementable algorithms for variational learning.
\end{abstract}


% Discussion August 8th
% ---
% Motivation for optimization over distributions
% - model the uncertainty in the cost function and the optimization procedure 
% - noise injection, additional stochasticity 
% - quality of the "solution", flat minimum
% - non-standard examples (e.g. PCA (stiefel manifold))
% - are we really solving the problem or is it just a way to come up with methods? 
% - change optimization to learning

\section{Introduction}
% Optimization problems over distributions are ubiquitous in machine learning but 
% can be difficult to solve due to constraints 
% on their parameters and the computation of low-variance gradients. 
Variational learning for neural networks~\citep{HiVC93,Gr11} offers many potential benefits over standard empirical risk minimization and can be used to motivate various deep-learning algorithms~\citep{KhRu23}. Some of the benefits of variational learning are improved regularization and reduced overfitting, as suggested by the minimum description length principle~\citep{HiVC93,Gr11,blier2018description,grunwald2007minimum} and PAC-Bayesian bounds~\citep{alquier2024user}. Depending on the form of the variational posterior, different deep learning algorithms can be obtained, for instance, Adam-like methods~\citep{ShDa24} for diagonal Gaussians, algorithms for binary neural networks for Bernoulli distributions~\citep{meng2020training} and biologically plausible multiplicative dynamics for log-normal distributions~\citep{nishidalog}.
The Bayesian viewpoint also offers a fruitful framework for thinking about model adaptation and model understanding, for example, federated and continual learning~\citep{nguyen2017variational,AsBu22}, model merging~\citep{matena2022merging,daheim2023model} or sensitivity analysis to perturbations in both the model and data~\citep{nickl2023memory}. 

%In the context of this paper, our main motivation is to design and study new learning rules for neural networks from Bayesian principles. %~\citep{KhRu23,KiKh23}.

% Recently proposed Lie group methods address the above difficulties but the relation 
% of such distribution learning methods to other algorithms is not fully understood.

The de-facto standard approach for variational learning in neural networks is to update the distribution parameters using gradient-based methods to leverage backpropagation training~\citep{Gr11}. \citet{KhRu23}, for example, use a natural gradient method on the natural parameters of an exponential family. This method has two main difficulties.
First, one needs to satisfy constraints on the parameters of 
the distribution, for example, positivity of the covariance matrix in the case of Gaussians~\citep{LiKh20}. Secondly, the objective functions involve integrals with respect to the variational distribution, and obtaining low-variance stochastic gradients is a major challenge. Reparametrization tricks provide such estimators, but it is not clear whether they are available for the variational family at hand and they have to be derived case by case~\citep{MoMn20}. 

Recently, \citet{KiKh23} proposed a learning algorithm which addresses the above difficulties by parametrizing the probability distribution through Lie groups. The constraints on the parameters are naturally handled through the group's closure and lower variance gradients are available by default as the Lie group automatically enables the reparametrization technique. Furthermore the homogeneity of the group allows transporting the gradients on the Lie group to a common tangent space thus allowing for a natural application of momentum, and such transports are compatible with the Fisher information metric which is invariant under reparametrizations.

In this work we establish the connection between the Lie group updates and existing methods on exponential families (EFs) such as the Bayesian Learning Rule (BLR) by \citet{KhRu23}.
The Lie group update rule does not require the family to be an EF, however many useful families are EFs. EFs are especially well suited to the Bayesian framework, which relies on multiplication and division of distributions to add, remove or modify information. This corresponds to simple addition and subtraction on the natural parameters of the family, connecting the Bayesian framework to optimizing on natural parameters. EFs  come with a rich structure, such as dually flat coordinate systems of natural and expectation parameters studied in information geometry~\citep{Am16}. As EFs have many desirable properties, the aim of this paper is to further study the Lie group learning rule of~\citet{KiKh23} when specialized to EFs and to understand its relation to the Bayesian learning rule~\citep{KhRu23} which is derived for EFs and gives a unifying and simplifying view of many other algorithms. 

Those EFs which also have a Lie group structure are families of distributions which combine the best of both worlds, and it should therefore not be surprising that most of the eponymous EFs named after a mathematician share this property. They have the above desirable properties of EFs, but the additional group structure gives easy reparametrization tricks and allows for updating without breaking constraints.
Developing a recipe to construct exponential families on homogeneous spaces from real representations of Lie groups, \citet{ToYo21} redefine the term \emph{harmonic exponential families} subsuming the earlier definition coined by \citet{cohen2015harmonic} for homogeneous spaces of compact Lie groups via unitary representations. These families characterize all exponential families closed under a Lie group action. 
Using this characterization we establish a general and concrete connection between the Lie group update and BLR. Our main contribution is to 
prove that the linearization of the update proposed by~\citet{KiKh23} always recovers 
the Bayesian learning rule for exponential families with Lie group structure, see~\Cref{thm:mainTheorem}. Altogether, our work supplies the theoretical foundations for efficient algorithms for variational learning using Lie groups and exponential families. 

%On top of the above contribution, by which we better understand BLR, we also study the following two examples:
%\begin{enumerate}
%	\item The multiplicative group, which gives rise to multiplicative updates for Bayesian learning. We study the relationship of these multiplicative updates to existing methods such as~\citet{LeSe00}, and we investigate the learning dynamics for training neural networks. 
%
%	\item The diagonal affine group provides families of learning algorithms which are similar to Varational Online Newton method~\citep{khan2018fast,KhRu23} which has recently been shown to be successful for large neural networks \citep{ShDa24}. The Lie group viewpoint suggests an additional momentum variable and a different retraction via the group exponential. We study the precise relationship between these two algorithms and carry out numerical comparisons.
%
%\end{enumerate}

\section{Background on Variational Learning}\label{sec:Background}
We first introduce some background on variational learning methods, with particular focus on exponential families. In the following section we introduce the Lie group method of \citet{KiKh23} to prepare for combining Lie groups with exponential families. 

\subsection{The Variational Learning Problem}
Given a parametrized model (for example, a neural network) with parameter space $\Theta$, choose a base measure $\nu$ on $\Theta$.
For a standard neural network the parameter space $\Theta$ is either $\R^P$ or an open subset of $\R^P$, and the base measure is generally taken as the Lebesgue measure.
Denote the loss function by $\ell : \Theta \to \R_{\geq 0}$. 
Instead of minimizing $\ell(\ttheta)$ to arrive at a $\hat{\vec{\theta}} \in \Theta$, variational learning obtains a distribution on the parameters via the optimization, 
\begin{equation}\label{eq:BLP}
\soln{q} \in \argmin_{q \in \mathcal{Q}} \mathcal{E}(q) \qquad \text{ where } \qquad \mathcal{E}(q) = \EE_{\text{$\ttheta$} \sim q}[\ell(\ttheta)] - \tau \mathcal{H}_{\nu}(q).
\end{equation}
Here $\mathcal{Q}$ is a variational family of distributions on $\Theta$ and $\tau > 0$ is a temperature parameter. Assume members of $\mathcal{Q}$ to be of the form $q\dd\nu$ for some density $q$, and by abuse of notation we often omit  $\dd\nu$ and simply refer to the distribution $q$---since the base measure will be fixed throughout, once chosen. We also assume that $\ell$ is $\nu$-almost everywhere differentiable and satisfies $\mathbb{E}_q[|\ell |]< \infty$ for every $q\in \Qcal$. The first is standard also in non-variational (point-based) learning if applying methods such as gradient descent, and the second assumption holds as long as members of $\Qcal$ have rapid decay and $\ell$ has moderate growth\footnote{In practice, since floating point numbers can only represent values in some interval, $[-M, M]$, we can force this condition by replacing $\ell$ with $\tilde{\ell}(\ttheta):= \min\{\max\{\ell(\ttheta), -M\}, M\}$ without any practical setback.}.

\begin{figure}[t!]
\includegraphics[width =\textwidth]{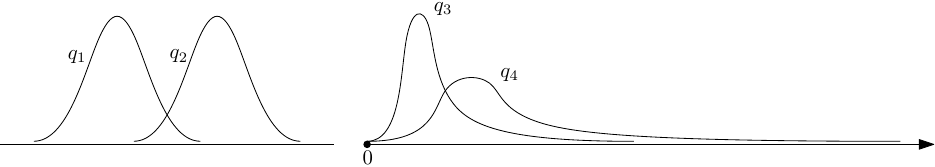}
 \caption{(Left) The distributions $q_1$ and $q_2$ are translates of each other and have the same entropy relative to the Lebesgue base measure $\dd\nu(\theta) = \dd\theta$. Therefore an algorithm optimizing for~\cref{eq:BLP} will not take entropy into account in deciding between them while finding $\soln{q}$. (Right) Two distributions on $\R_{>0}$ are related by $q_4(\theta) = \tfrac{1}{2}q_3(\tfrac\theta 2)$. Relative to the Lebesgue measure, $q_4$ is spread out more and has a higher entropy, however both measures have the same entropy relative to the base measure $\dd\theta/\theta$. In this case the densities are given by $\tilde{q}_j$ with $\tilde{q}_j(\theta) = \theta q_j(\theta)$ ($j = 3,4$) and $\tilde{q}_4(\theta) = \tilde{q}_3(\tfrac \theta 2)$. Therefore the solution to the Bayesian Learning problem in \cref{eq:BLP} depends on the base measure and it is the practitioner's choice whether to bias the solutions $\soln{q}$ towards $q_4$ ($\dd\nu(\theta) = \dd\theta$) or not ($\dd\nu(\theta) = \tfrac{\dd\theta}{\theta}$).}\label{fig:entropy_base_measure}
\end{figure}

The term $\mathcal{H}_\nu(q) = - \int_{\Theta} q \log q \dd\nu$ denotes the entropy of the measure $q\dd\nu$ relative to the base measure $\nu$. The base measure underlies the concept of entropy, see~\Cref{fig:entropy_base_measure} for how the choice of base measure can affect the solution. Indeed, if we have two ways of writing the same measure $q_1\dd\nu = q_2 (h \dd\nu)$ their relative entropies are not the same and are related by the equality $\mathcal{H}_\nu(q_1) + \EE_{q_1\dd\nu}[\log h] = \mathcal{H}_{h\nu}(q_2)$.  We introduce the required notions from measure theory later when we need them in~\Cref{ssec:notation}. 

The notion of relative entropy is exactly the $\DKL$-divergence $\mathcal{H}_\nu(q) = -\DKL(q\dd\nu \|\nu)$ if $\nu$ is also a probability measure. The variational learning problem is sometimes (equivalently) introduced as $\mathcal{E}(q) = \EE_{q}[\ell_{\text{data}}] + \tau \DKL(q\| p_{\text{prior}})$, where the prior is chosen as $p_{\text{prior}} \propto e^{-\frac{1}{\tau}\ell_{\text{reg}}} \dd\nu$ following a decomposition of the loss $\ell = \ell_{\text{data}} + \ell_{\text{reg}}$ into the data term and regularizer.

In~\Cref{eq:BLP}, the expectation term $\EE_q[\ell]$ pushes the optimization towards $q$ that are more concentrated in regions of low loss. Antagonistically the negative entropy term $-\mathcal{H}_{\nu}(q)$ pushes the optimization towards $q$ which are more spread out, exploring the parameter space. The objective $\mathcal{E}$ can thus be seen as an implementation of the exploration--exploitation trade-off in the parameter space. The temperature $\tau$ balances this trade-off. The entropy term is an additional regularization on top of what might already exist in $\ell$, selecting for models robust to larger parameter noise. These ideas have been studied from the minimum description length principle perspective~\citep{HiVC93,Gr11,blier2018description,grunwald2007minimum} and through PAC-Bayesian bounds~\citep{alquier2024user}. 

%Having access to a whole distribution $\soln{q}(\ttheta)$ rather than a single $\soln{\ttheta}$ is useful in formulating methods of model adaptation and model understanding, for example, federated and continual learning~\citep{nguyen2017variational,AsBu22}, model merging~\citep{matena2022merging,daheim2023model} or sensitivity analysis to perturbations in both the model and data~\citep{nickl2023memory}. 

If $\ell(\ttheta)$ is the sum of negative log-likelihood and the negative log-prior, then~\cref{eq:BLP} is equivalent (with $\tau = 1$) to minimizing the Kullback-Leibler divergence $\argmin_{q \in \Qcal} \DKL(q\|p)$ from the ``Gibbs'' posterior $p \propto p(\mathcal{D}\,|\,\vparam) p_{\text{prior}}(\ttheta) \propto e^{-\ell(\vec{\theta})}$, leading to a Bayesian interpretation \citep{Ze88}. A general loss function $\ell$ leads to generalized posteriors,  \citep{Zh99,Ca07,KnJe19, BiHo16}. 

The set $\mathcal{Q}$ over which we restrict \Cref{eq:BLP} is also called the space of candidate posteriors. An important family of candidate posteriors $\mathcal{Q}$ are exponential families, which have many desirable properties. We introduce them in the following subsection, and subsequently the BLR algorithm solving for \Cref{eq:BLP} with exponential families.

\subsection{Exponential Families}

First construct an exponential family of distributions on $\Theta$. For this purpose, fix a pair $(\nu, \vT)$ where $\nu$ is a base measure on $\Theta$ and $\vT: \Theta \to \R^d$ is a measurable function\footnote{Here $\Theta$ is a topological space (in fact a manifold) and thus given its Borel sigma algebra, we assume $\vT$ is measurable and $\nu$ is a Radon measure.} with coordinate functions  $\vT(\ttheta) = (T_1(\ttheta), T_2(\ttheta), \ldots, T_d(\ttheta))^\top$. This $\vT$ is called a \emph{sufficient statistic}.
Given $\vT$ and $\nu$ as above, we construct the associated (maximal) exponential family as 
\begin{equation} \label{eq:ExponentialFamily}
	\EF_{\nu}(\vT) = \left\{ q_\llambda \dd\nu : q_\llambda(\ttheta) =  \exp\left(-\llambda^\top \vT(\ttheta) - A(\llambda) \right): \llambda \in \Omega \right \},
\end{equation}
where $A=A_{\text{$\nu, \vT$}} : \R^d \to (-\infty, \infty]$ is the log partition function defined as
\begin{equation}\label{eq:logPartitionDefinition}
	A(\llambda) = \log \int_{\Theta} \exp\left (- \llambda^\top  \vT(\ttheta)\right) \dd\nu(\ttheta) 
\end{equation}
and $ \Omega_{\nu,\text{\vT}} = \{\llambda \in \R^d : A(\llambda)< \infty\}$ is the effective domain of $A$. The log-partition function $A(\llambda)$ is convex and therefore the set $\Omega = \Omega_{\nu, \vT}$ is convex as well.

The vectors $\llambda$ are called the \emph{natural parameters} of the distribution $q_{\llambda} \dd\nu$. The coordinate-free notation is also revealing; consider $\vT : \Theta \to \primsp$ where $\primsp$ is a finite dimensional real vector space\footnote{which is isomorphic to $\R^d$ after a choice of basis} and in this case $\llambda$ is naturally an element of the dual space $\dualsp$.
%, whose role in~\cref{eq:ExponentialFamily} is to send the vector $\vT(\ttheta)$ to real numbers. 
This coordinate-free notation lets us distinguish between the sufficient statistics and natural parameters which are otherwise both simply elements of $\R^d$. The linear dual space---denoted by $\dualsp$---is the set of real valued linear functionals on $\primsp$, which itself forms a vector space. The value of the linear functional is denoted by the inner product notation $\llambda(\vv) = \langle \llambda, \vv\rangle$ for $\vv\in \primsp$ and $\llambda \in \dualsp$. This coincides with the coordinate-laden description above; upon choosing a basis of $\primsp$ and the corresponding dual basis in $\dualsp$, both $\vT(\vec{\theta})$ and $\llambda$ can be written as column vectors and the pairing is given by the dot product $\langle \llambda, \vT(\ttheta) \rangle = \llambda^\top \vT(\ttheta)$.

The discussion above is a constructive approach to exponential families. Alternatively, we may as in~\Cref{app:ExponentialFamily}, from~\citep[\S~8.1; Def.~2.2]{bn78, ToYo21}, give criteria dubbing a set $\mathcal{P}$ of probability measures as an exponential family or not. Their approach does not commit to a base measure or sufficient statistics, as the \emph{same} set of probability distributions may be realized as an exponential family with two (or more) realizations given by $R = (\nu, \vT)$, $R' = (\nu', \vT')$. It also allows for considering smaller families than all normalizable ones. Given a realization $R$ of an exponential family $\mathcal{P}$, denote the set of natural parameters corresponding to  the family as $\Omega_{R}(\mathcal{P}) \subseteq \Omega_R \subseteq \dualsp$. We drop $R$ from the notation if it is understood. It is worth going into some delicate technicalities about different realizations of a set of probability measures as an exponential family.
\begin{example}\label{ex:1dgaussian}
Take $\mathcal{P}$ as the one-dimensional Gaussian distributions with fixed variance $1$. This set of distributions can be realized as an exponential family in two ways.

Take $\dd\nu(x) = \dd x$, i.e.\ the Lebesgue measure, and $\vT(x) = (x, \tfrac12 x^2)$ as the sufficient statistic $\vT : \R \to \R^2 \cong \primsp$. In this case the distribution $p =\tfrac{1}{\sqrt{2\pi}} \exp( -\tfrac12 (x- m)^2)\dd x \in \mathcal{P}$ can be associated with the natural parameter $\llambda = (-m, 1) \in \R^2 \cong\dualsp$. This natural parameter space $\Omega_{\dd x, \text{\vT}} (\mathcal{P})$ lies inside the set of all feasible natural parameters $\Omega_{\nu, \text{\vT}} = \{(\lambda_1, \lambda_2) \in \R^2: \lambda_2 > 0\}$ which constructs the space of all 1-d Gaussians.

As a second realization we can take $\dd \nu'(x) = e^{-\frac{x^2}{2}} \dd x$ as the base measure and $\vT' : \R\to \R\cong \primsp$, with $\vT'(x) = x$. In this case $\Omega_{\nu', \text{\vT}'}(\mathcal{P}) = \R \cong \dualsp$. In this second realization $\mathcal{P}$ is a full exponential family (given $\nu'$ and $\vT'$) and furthermore it is a minimal realization of the exponential family. 
\end{example}

The distributions $q_{\llambda} \dd\nu$ in $\EF_\nu(\vec{T})$ can be characterized as 
\[
	\argmax_{q}~\mathcal{H}_\nu(q) \qquad \text{ subject to }\qquad  \EE_{q\dd\nu}[\vT] = \mmu,
\] 
where $\mmu \in V$ is called the \emph{mean} or the \emph{expectation} parameter\footnote{This characterization connects the exponential family with variational Bayesian learning, solving for maximum entropy under the constraint $\EE_{q}[T_i] = \mu_i$ leads to the same kind of optimization as~\cref{eq:BLP}.}. To be precise, given $\vT$, $\nu$ and $\mmu$, if there is a $\llambda \in \Omega_{\nu,\text{\vT}}$ such that $\EE_{q_\llambda}[\vT]= \mmu$ then any other distribution satisfying the same constraint has lower entropy. The proof in~\citet[Theorem 12.1.1]{cover2009} extends to the general base-measure setting. Thus expectation parameters characterize elements of $\mathcal{P}$, each $\mmu$ giving at most one distribution in $\mathcal{P}$. This shows the conceptual centrality of exponential families, as they can be interpreted as implementing the principle of indifference (principle of maximum entropy) given the expectation information.

There are two canonical coordinatizations of $\mathcal{P}$, the \emph{natural parameters} $\llambda \in \Omega(\mathcal{P}) \subseteq \dualsp$ and the \emph{expectation parameters} $\mmu \in \primsp$. If the functions $1, T_1(\ttheta), \ldots, T_d(\ttheta)$ are not linearly independent then multiple $\llambda$ correspond to the same distribution. This can easily be fixed by omitting certain coordinates from $\vT$, and let us assume that we have done that. Also not every choice of $\mmu \in \primsp$ gives a valid distribution, for example the $1$-d Gaussians (with non-fixed variance), have expectation parameters $\mmu = (m, \tfrac12 (m^2 + \sigma^2)) = \EE_{x \sim \mathcal{N}(m, \sigma^2)}[(x, \tfrac12 x^2)]$. So only those $\mmu\in \primsp$ lying in the (strict) epigraph of the parabola $y = \tfrac12  x^2$ are feasible expectation parameters corresponding to a natural parameter $\llambda$. If $\llambda$ lies in the interior of $\Omega_{\nu, \text{\vT}}$ then the gradient of the log-partition function connects the two parametrizations as $\mmu(\llambda) = \EE_{q_\llambda}[\vT] = -\nabla A(\llambda)$,  which can be seen via direct calculation. A full exponential family $\mathcal{P}$ with a minimal realization $R = (\nu, \vT)$ is called \emph{regular} if its set of natural parameters $\Omega_{R}(\mathcal{P})$ is an open subset of $\dualsp$, in this case $-\nabla_\llambda A$ gives the change of variables formula between the two coordinates. We refer the interested reader to \citet[Chapter~2]{Am16} and \citet[Chapter~3]{WaJo08}.

%The exponential family is called minimal if the image of $\vT$ does not lie in a proper affine subspace of $V$ and $\Omega_\nu(\vT)$ does not lie in a proper affine subspace of $\dualsp$. This prevents cases $\llambda \neq \llambda' $ describing the same distribution. For minimal exponential families $\nabla A$ is injective. We will assume minimality in the rest of our paper for convenience. 

% EFs are especially well suited to the Bayesian framework, which relies on multiplication and division of distributions  Multiplication and division of EFs simplifies to addition and subtraction on the natural parameters of the family, which connects the Bayesian framework to optimization techniques used in machine learning such as gradient descent, which are based on addition. EFs are maximum entropy distributions and come with a rich structure, such as dually flat coordinate systems of natural and expectation parameters studied in information geometry~\citep{Am16}.\footnote{Duality has been studied earlier than 2016. Old Amari's paper should be cited. \cmehmet{Yes, Please cite the paper you are referring to as well} }

If we select the candidate posteriors $\Qcal$ in~\cref{eq:BLP} as an exponential family then updating the distribution is the same as updating the natural parameter $\llambda$. In the following section we discuss a practical gradient-based algorithm updating $\llambda$, the BLR.

%\subsection{\ckoichi{About the definition of regularity and minimality in ``Fast and Simple Natural-Gradient Variational Inference with Mixture of Exponential family''( this is not a original subsection, just for an explanation of notion of exponential family)}
%}
%\ckoichi{
%In this paper, 
%``regular exponential family'' is introduced first and after that ``A regular EF representation is said to be minimal'' is defined. 
%However, regularity is not well-defined without the minimality in my opinion. 
%For example, the family of normal distributions on $\R$ can be written as follows:
%\begin{align*}
%  &\left\{ \exp(- \pairing{ \lambda}{ \begin{pmatrix}\theta\\ \theta^2 \end{pmatrix}}-A(\lambda))d\theta \right\}_{\lambda\in \Omega}\\
%  &=\left\{\exp(-\pairing{\lambda'}{\begin{pmatrix} \theta \\ \theta^2 \\ \theta^3\end{pmatrix}} -A'(\lambda')d\theta\right\}_{\lambda' \in \Omega'},\\
%        \Omega&=\R \times \R_{>0}\subset \R^2,\\
%        \Omega'&=\R\times \R_{>0}\times \{0\}\subset \R^3. 
%\end{align*}
%$\Omega$ is open in $\R^2$ but $\Omega'$ is not open in $\R^3$
%Therefore, we can not determine whether the family is normal distributions is regular or not. 
%However, if we  consider only the minimal representations in the sense of Definition~\ref{def:exponential_family}, whether the parameter space is open or not does not depend on the choice of the minimal representation. 
%Thus, we can define the concept of regular exponential family intrinsically. 
%}

\subsection{Variational Learning in Exponential Families}\label{subsec:variationalLearning}
The Bayesian Learning Rule (BLR) by~\citet{KhRu23} is variational learning on an exponential family, in which $\mathcal{E}$ is minimized by Natural Gradient Descent (NGD) on $\llambda$. NGD means that the Fisher information metric is chosen as the Riemannian geometry on $\Omega(\mathcal{P})$ and the natural parameters are updated using Riemannian gradient descent.

Selecting an open exponential family as candidate posteriors, we minimize $\mathcal{E}(\llambda) := \mathcal{E}(q_{\llambda})$ with gradient descent on $\llambda$. Mention of gradients assumes a choice of metric, and if none is specified, it is implicitly the Euclidean geometry. Given a different geometry such as the Fisher information metric, the corresponding gradient is denoted by $\widetilde{\nabla}_\llambda \mathcal{E}$, called the natural gradient. This vector is concretely computed by preconditioning the Euclidean gradient with the Fisher information matrix $F_\llambda$.

Altogether, this leads to the following update, 
\begin{equation}\label{eq:BLR1}
	\llambda^{\text{updated}} = \llambda - \alpha \widetilde{\nabla}_\llambda \mathcal{E}(\llambda) = \llambda - \alpha \vF_{\text{$\llambda$}}\inv \dd \mathcal{E}\big|_\llambda, 
\end{equation}
for step size $\alpha>0$. Here $\dd \mathcal{E}\big|_{\llambda}$ refers to the differential of $\mathcal{E}$ which naturally lives in the cotangent space $T_{\llambda}^* \Omega(\mathcal{P}) \cong (\dualsp)\dual \cong \primsp$.  In the basis $\{\dd\lambda_i\}_i$ this differential can be written as $\nabla_\llambda \mathcal{E}(\llambda)$, a vector with partial derivatives $\tfrac{\partial \mathcal{E}}{\partial \lambda_i}$ as coordinate values, and in coordinates the natural gradient is $\widetilde{\nabla}_\llambda \mathcal{E}(\llambda) = \vF_\llambda\inv \nabla_\llambda \mathcal{E}(\llambda)$.  Geometrically speaking, the natural gradient is a tangent vector in $T_\llambda\Omega(\mathcal{P}) \cong \dualsp$, and the inverse Fisher matrix can naturally be seen as a transformation mapping the differential co-tangent vector back to the tangent space. More details on this geometric point of view can be found in~\Cref{app:subsecGradvsDiff}.

If the base measure $\nu$ in~\cref{eq:BLP} coincides with that of the exponential family's realization $\dd \mathcal{H_\nu}\big|_{\llambda}(q_\llambda) = -\vF_\llambda \llambda$, then we get for $\tau = 1$, 
\begin{equation}
  \llambda^{\text{updated}} = (1- \alpha) \llambda + \alpha \vF_\llambda\inv \dd \EE_{q_\llambda}[-\ell(\vparam)]\big|_\llambda,
	\label{eq:BLR2}
\end{equation}
cf. BLR in~\citet[Eq.~3]{KhRu23}. Implementing the update in~\Cref{eq:BLR2} comes with two difficulties. First, if the learning rate $\alpha$ is large the update may overshoot and leave the constraint set of admissible parameters $\vlambda \in \Omega$. Secondly, the formalism of exponential families by itself does not provide enough structure for efficiently calculating a low-variance stochastic approximation to the gradient of the expected loss. Where such calculation is tractable, we observe that a reparametrization trick is available because a Lie group acts on the parameters and the distributions. In the following section, we introduce the Lie Group Bayesian Learning Rule (LG-BLR) of \citet{KiKh23}, which addresses these issues.

%the Bayesian Learning Rule (BLR) of \cite{KhRu23} was developed, given as a Riemannian gradient descent on the $\llambda \in \dualsp$ using the Fisher information metric, which is also called the Natural Gradient Descent (NGD). In order to make this precise let use rewrite \eqref{eq:BLP} as
%\begin{equation}\label{eq:BLPviaKL}
%	q^* \in \argmin_{q \dd\nu \in \Qcal} \DKL(q \| p_\tau) \qquad \text{ where } p_\tau(\ttheta) \propto e^{-\frac{1}{\tau} \ell(\ttheta)}
%\end{equation}
%\cthomas{We have to discuss whether there is a better way than having the $q \dd \nu$ notation.} 

%Here the Kullback-Leibler divergence of measures is $\DKL(q\|p) = \int_{\Theta} q \log \frac{q}{p} \dd\nu.$  We can call $\mathcal{E}(q) := \DKL( q\|p_\tau)$ for convenience. Notice that $\mathcal{E}(q) = \EE_{q}[\tfrac{1}{\tau} \ell] - \mathcal{H}_\nu(q) + \text{const.}$ and so the argmin is not affected. 

%The Fisher information metric is defined on any parametrized family as a matrix (a bilinear form on the tangent space after choosing basis vectors) with entries
%\[
%	[F_\llambda]_{i,j} = \EE_{q_\llambda \dd\nu} \left[\left( \tfrac{\partial}{\partial \lambda_i} \log q_\llambda \right) \left(\tfrac{\partial }{\partial \lambda_j} \log q_\llambda \right)\right]
%\]

\section{The Lie Group Method of \citet{KiKh23}} \label{sec:LGBLR}

Estimating the derivative of the expected loss in~\Cref{eq:BLR2} has garnered much interest, see \citet{MoMn20} for a survey of various methods. Pathwise gradient estimators are one low-variance stochastic estimation method. This technique requires that the distributions in the variational family come from differentiable reparametrizations of a base \emph{noise}, such as in the case of Gaussian distributions. Upon translations and scaling, the stochastic aspect of sampling from any Gaussian distribution can be relegated to  sampling noise vectors from $\mathcal{N}(\vparam \mid \mathbf{0}, I)$. This generalizes to all similar cases where a Lie group of transformations acts differentiably on $\Theta$ and the distributions in $\Qcal$ are pushforwards of each other via these transformations. In many cases using the reparametrization method reduces the variance in stochastic estimation of the gradient of expectation in comparison with methods such as score function estimator, see~\citet{Xu19}. This is the setting of LG-BLR~\citep{KiKh23}, which is a learning algorithm for the variational problem in~\cref{eq:BLP}. The purpose of this work is to show that  in cases such as families of Gaussian or exponential distributions, where $\Qcal$ is both an exponential family and the Lie group setting applies, the linearization of the LG-BLR yields the BLR. This gives insights on BLR itself, which we mention later. 

\subsection{Notation for the Lie Group Update}
\label{ssec:notation}

The setup of \cite{KiKh23} involves the notions of Lie groups, group actions, homogeneous spaces, and relatively invariant measures. For details about groups in the machine learning context \cite[pp. 13, 40]{Br21}. For an introduction to measure theory in machine learning context, see the blog post~\citep{MeasureBlog24}. Below is a short summary.

A \emph{group} is a set $G$ with a binary operation $*$ satisfying the properties of associativity, existence of an identity element and inverses. These mean: first, $g*(h*k) = (g*h)*k$ for all $g,h,k \in G$; second, there exists an identity element $e \in G$ such that $e*g = g*e = g$ for all $g\in G$; finally, for any $g\in G$ there exists an inverse element which we denote by $g\inv \in G$ such that $g*g\inv = g\inv * g = e$.
A \emph{Lie group} is also a smooth manifold, and both its binary group-operation and inversion are smooth. 
For groups written with a multiplicative notation, it is common to write $gh$ in place of $g*h$. A smooth manifold is locally diffeomorphic to Euclidean space, that is, there are infinitely differentiable invertible mappings between local patches of $G$ and an open subset of $\R^m$, and $m$ is called the dimension of the manifold. As an example the set of invertible matrices on $n$-dimensions is a group with respect to matrix multiplication, is a Lie group of dimension $n^2$.

The tangent space of the Lie group at the identity is called a Lie algebra, denoted as $\mathfrak{g} = T_eG$. It is a vector space, which also has a binary operation called \emph{the Lie bracket} denoted by $[X, Y] \in \mathfrak{g}$ for $X,Y \in \mathfrak{g}$. For matrix groups (i.e. subgroups of invertible $n\times n$ matrices) the Lie algebra is a vector space of square matrices and the Lie bracket is concretely given as $[X,Y] = XY - YX$. 

The differential of the left-multiplication-by-$g$ map $L_g: x \mapsto gx$ on $G$ provides a consistent way to transfer a direction vector $v\in \mathfrak{g}$ to a tangent vector $\dd L_g\big|_e(v) \in T_gG$. In LG-BLR we make use of this, ensuring that all the gradient vectors end up in the same vector space $\mathfrak{g}$, so that we can accumulate past gradients via momentum\footnote{In standard gradient descent (GD) on $\R^n$ gradients at different points $\vx_0$ and $\vx_1$ also live in different tangent spaces $T_{\vx_0}\R^n, T_{\vx_1}\R^n$. The Lie group of translations acts transitively on the domain, and the differential of the map $\vx \mapsto \vx + (\vx_1 - \vx_0)$ identifies these tangent spaces. This is all implicit in concrete computation. Such an identification makes GD acceleration methods (e.g.\ momentum) meaningful.}.

The right-multiplication-by-$g$ map $R_g : x \mapsto xg$ has the same effect, which is different from $L_g$ if the group multiplication is not commutative. The difference between the left and right multiplication maps can be measured by the conjugation automorphisms, which for a given group element $g \in G$ is the map $x \mapsto gxg\inv$. It is differentiable and $e\mapsto e$, so its differential at identity is a linear map $\operatorname{Ad}_g : \mathfrak{g} \to \mathfrak{g}$. It is called the \emph{adjoint} map. For commutative groups $G$, this map is always the identity. For matrix groups, $g \in G$ and $X \in \mathfrak{g}$ are both square matrices and $\operatorname{Ad}_g(X) = gXg\inv \in \mathfrak{g}$.

Every Lie group comes with the structure of an \emph{exponential map}, allowing us to wrap the tangent space at identity to the group manifold. The exponential $\exp_{G}: \mathfrak{g} \to G$ is a map defining paths---also called one parameter subgroups---in $G$ by $\{\gamma_X(t) := \exp(tX) : t\in \R\}$. These paths are closed under the group multiplcation and the direction at $t = 0$ is $\gamma_X'(0) = X$. These are geodesics\footnote{with respect to the canonical left invariant affine connection where parallel transports are given by $\dd L_g$.} on $G$ starting at the identity $e$, with initial velocity $X$. If the group is understood we may simply write $\exp_G(X) = \exp(X) = e^{X}$, dropping $G$ from the notation. Given two directions $X, Y \in \mathfrak{g}$ such that $[X,Y] = 0$, the exponential map is multiplicative; $e^{X + Y} = e^X e^Y$. In particular $e^{(t + s)X} = e^{tX} e^{sX}$, justifying why we call $\gamma_X$ as \emph{one parameter subgroups}.
For matrix groups $G$, the Lie algebra $\mathfrak{g}$ is a subspace of square matrices and the exponential map is given by $\exp(X)= \sum_{n = 0}^\infty \frac{X^n}{n!}$. 

A \emph{group action} on a set $A$ is a map $G \times A \to A$ denoted by $g\cdot a$ satisfying $(gh)\cdot a = g\cdot (h\cdot a)$ and $e\cdot a = a$ for any $g,h\in G$, $a \in A$ and $e$ as the identity element of $G$. The set $\{g\cdot a: g\in G\} \subseteq A$ is called the \emph{orbit} of $a$. A group action is \emph{transitive} if there is some element $a_0 \in A$ such that any $a \in A$ is in the orbit of $a_0$. %In such a case the basis point $x_0$ is not special and any other point can also be chosen as a basis point. 
The stabilizer of an element $a_0$ is the subgroup of $G$ fixing $a_0$, denoted by $\operatorname{Stab}_G(a_0) = \{g\in G: g\cdot a_0 = a_0\}$. If transitive, the set $A$ and the quotient set $G/H$ (where we call $H = \operatorname{Stab_G}(a_0)$) can be identified by associating elements $a \in A$ with the group elements $g$ that take a basis point $a_0$ to $a$ (modulo the group elements that fix $a_0$).  If $A$ is a manifold, with such a transitive (and continuous) $G$-action, $A$ is called a \emph{homogeneous space}. Furthermore if $A$ is a differentiable manifold with a smooth group action, differentiating the group action at the identity gives the \emph{infinitesimal group action}, which we denote with the same symbol: $X \cdot a := \frac{\dd}{\dd t} (e^{tX} \cdot a)\big|_{t = 0}$. 

A \emph{representation} of a group $G$ is a group homomorphism $\pi : G \to \GL(\primsp)$ for some vector space $\primsp$. Given a basis of $\primsp$ this means associating matrices $\pi(g)$ to elements of the group such that group operation corresponds to matrix multiplication: $\pi(g_1g_2) = \pi(g_1)\pi(g_2)$. This makes abstract groups concrete, since matrices are concrete objects. A faithful representation for example, i.e. one where $\pi$ is injective, replicates the group structure of $G$ inside some $n\times n$ invertible matrices, where $n = \dim \primsp$. Given a group representation $\pi$ on $\primsp$, there is a dual representation $\pi^* : G \to \GL(\dualsp)$. The action of $\pi^*(g)$ on a linear functional $\llambda \in \dualsp$ is defined by $\langle \pi^*(g)\llambda, \vv \rangle = \langle\llambda, \pi(g\inv)\vv \rangle$ for every $\vv \in \primsp$.  As matrices (in the dual basis of $\dualsp$) $\pi^*(g)$ is given by the inverse transpose of $\pi(g)$. 

A \emph{measure} $\nu$ on $A$ assigns a nonnegative number to members of a certain collection of subsets (called a sigma-algebra\footnote{set of subsets of $A$ closed under countable unions, complements and containing $A$} elements of which in this context are called measurable sets) with countable additivity and $\nu(\emptyset) = 0$. For example the Lebesgue measure assigns an interval $(a, b)$ its length $b - a$. Given a map $\phi:A \to B$ there is a measure $\phi_\sharp \nu$ on $B$ called the \emph{pushforward measure} such that $(\phi_\sharp\nu)(E) = \nu(\phi\inv(E))$ for all measurable sets $E$. For this $\phi$ needs to be ``measurable'' which all continuous functions are if $A$ also has a topology and the sigma-algebra is constructed from the open sets of $A$. 

Assume the measure space $A$ also has a measurable group action on it, then for each $g\in G$, the map $a \mapsto g\cdot a$, yields the pushforward measure $g_\sharp\nu$. Assume further that $\nu(g E) = \chi(g) \nu(E)$ for some factor $\chi(g)$ not depending on the measurable subset $E$, then we have $g_\sharp\nu = \frac{1}{\chi(g)}\nu$. In this case $\nu$ is called a \emph{relatively $G$-invariant} measure. Such a measure is not guaranteed to exist, but in all important cases we present, it exists and can be explicitly written down. The multiplier $\chi : G \to \R_{>0}$  is necessarily a group homomorphism, i.e. it satisfies $\chi(g_1g_2) = \chi(g_1) \chi(g_2)$ for all $g_1, g_2 \in G$. We also introduce a bit of notation which is necessary later, and define the differential of $-\log\chi$ as $\xi$, i.e. $\xi: \mathfrak{g} \to \R$ defined by $\xi(X) := -\tfrac{\dd}{\dd t} \log (\chi(\exp(tX)))\big|_{t = 0}$.

  As an example, take $A = \R^P$ and $\nu$ the Lebesgue measure, and $G = \operatorname{Aff}_P(\R)$ with the action given by $(A, \mathbf{b})\cdot \mathbf{x} = A\mathbf{x} + \mathbf{b}$. Then $(A,\mathbf{b})_\sharp \nu = |\det(A)|^{-1} \nu$  where the factor $\chi((A,\mathbf{b})) = |\det(A)|$ is the Jacobian determinant coming from the change of variables formula. The Lie algebra $\mathfrak{aff}_P$ consists of pairs $(X, \vec{x})$ with all $P\times P$ matrices $X$ and $\vec{x} \in \R^P$. In this case $\xi((X, \vec{x})) = - \frac{\dd}{\dd t} \log(\det(e^{tX}))\big|_{t = 0} = - \tfrac{\dd}{\dd t} \Tr(tX) \big|_{t = 0} = -\Tr(X)$.

In terms of change of variables this relative--$G$-invariance property of $\nu$ means that for any measurable function $f$ on $A$ we can write
\[
	\int_{A} f(a) \dd\nu(a) = \int_{A} f(g\inv \cdot a) \dd (g_\sharp\nu)(a)  
 = \frac{1}{\chi(g)} \int_{A} f(g\inv \cdot a) \dd\nu(a). 
 \]
This is very useful for calculating gradients of loss expectations in our LG-BLR setting.
%In case $A = \R^P$ and $G =\Aff_P(\R)$ is the affine group acting on it, then a relatively $G$-invariant measure $\nu$ has to equal the Lebesgue measure up to positive scalar. 

\subsection{The Lie Group Update Rule} \label{sec:LieGroupBLR}

Assume the setting above, i.e. let $G$ be a Lie group acting smoothly and transitively on the space of parameters $\Theta$ of a machine learning model, and $\nu$ is a relatively invariant measure on $\Theta$ with multiplier $\chi$. Further assume $\Theta$ is a differentiable manifold, and without loss of generality we may assume it is embedded in some ambient Euclidean space $\R^n$. Fix a base distribution $q_0\dd\nu$. The group action map $\ttheta \mapsto g\cdot \ttheta$ induces a pushforward on the distributions given as  $g_\sharp(q_0\dd\nu) = q^g \dd\nu$ where
\[
q^g(\ttheta) = \frac{1}{\chi(g)}q_0(g\inv \ttheta)
\]
is the density of the pushforward distribution\footnote{This notation is different from the one used in \cite{KiKh23} but in order to avoid a notation clash with natural parameters, we denote the pushforward density by a superscript instead of a subscript.}. Sampling $\ttheta$ from $q^g\dd\nu$ is equivalent to sampling $\ttheta_0 \sim q_0\dd\nu$ and applying the group action $\ttheta = g \cdot \ttheta_0$. 

 In LG-BLR of~\citet{KiKh23}, the family of candidate posteriors is chosen as 
 \begin{equation}\label{eq:Qcaldefn}
 \Qcal = \{q^g \dd\nu : g\in G\}.
 \end{equation} Members of $\Qcal$ are parametrized by Lie group elements, and optimizer steps update $g \in G$.

\begin{figure}[h]
\centering
\includegraphics[width =\textwidth]{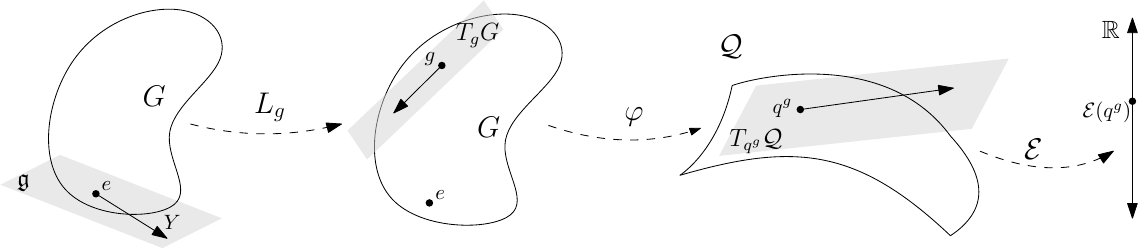}

\caption{The direction $Y \in \mathfrak{g}$ is carried to $T_gG $ by the differential of $L_g$ and that is carried to the tangent space $T_{q^g}\mathcal{Q}$ by $\dd\varphi\big|_{g}$. We solve for $Y = \hat{Y}$ so that this tangent vector on $\Qcal$ is in the direction of fastest ascent according to the Fisher metric.}\label{fig:groupComposition}
\end{figure} 
 
 For a step size $\alpha > 0$, the update is given as 
\begin{equation} \label{eq:gupdate}
	g^\text{updated} = g \exp(-\alpha \hat{Y})
\end{equation}
where $\hat{Y} \in \mathfrak{g}$ is in the direction\footnote{As mentioned previously, $\mathfrak{g}$ denotes the Lie algebra of the group $G$. The updates are structured as to update by multiplying with ``small'' (i.e. close to identity) elements along the \emph{directions} $\hat{Y}$ at identity.} of fastest ascent for the objective $\mathcal{E}(q)$ at $q = q^g$. The notion of \emph{fastest ascent} requires specifying a metric, 
and in accordance with the original BLR \citep{KhRu23} we choose the Fisher information metric. To be precise, let $B$ be the pullback of the Fisher information metric on $\Qcal$ to $\mathfrak{g}$, by $\varphi \circ L_g$, see \cref{eq:FisherBilinearOnGroup} in \cref{app:liegroupblr}. Here $L_g : G\to G$ is the left-multiplication-by-$g$ map and $\varphi : G \to \Qcal$ is the 
parametrization $\varphi(g) = q^g$ from the group to the densities. As the update direction choose a $\hat{Y} \in \mathfrak{g}$ satisfying 
\begin{equation} \label{eq:FastestAscentOptimization}
B(\hat{Y}, X) = \dd(\mathcal{E} \circ \varphi \circ L_g)\big|_{e}[X] \qquad \text{ for all } \quad X \in \mathfrak{g}.
\end{equation}
One can show $\hat{Y} = s \hat{X}$, where $s = \dd (\mathcal{E} \circ \varphi \circ L_g)\big|_{e}[\hat{X}]$ and where $\hat{X}$ solves
\begin{equation*}
\hat{X} \in \argmax_{X\in \mathfrak{g}}  ~  \dd(\mathcal{E} \circ \varphi \circ L_g)\big|_e [X]  \qquad \text{such that}\quad  B(X, X) \leq 1
\end{equation*}
So, in words, $\hat{X}$ is the direction of fastest ascent according to $B$, and $s$ is the maximal slope. This intuitive picture is valid even if our $B$ has degenerate directions. The existence of $\hat{Y}$ requires that the differential vanish on such directions, which it does. 

\begin{figure}[t!]
\centering
\includegraphics[width =0.56\textwidth]{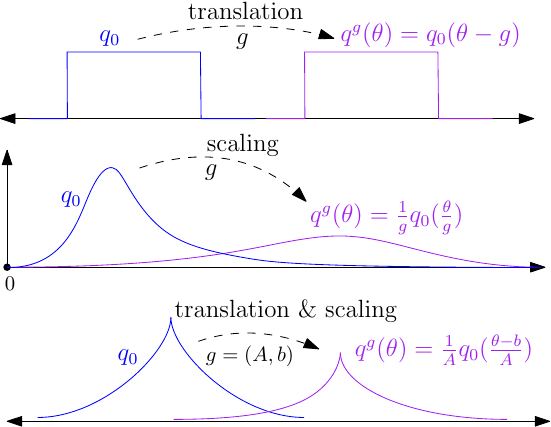}
\includegraphics[width = 0.43\textwidth]{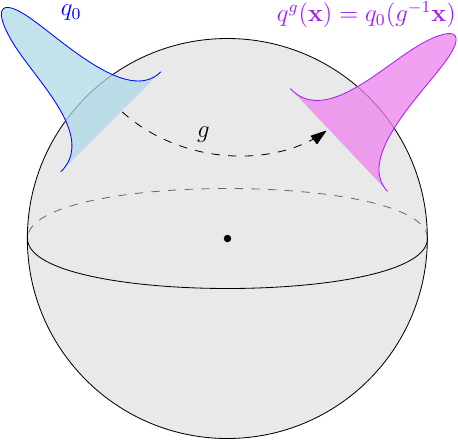}

\caption{Examples of Lie groups acting on $\Theta$ and distributions on $\Theta$. (top-left) $\Theta = \R$ and $G = (\R, +)$, with action given by $g \cdot \theta = g + \theta$. The orbit of the pushforwards are various translations of $q_0$. (middle-left) $\Theta = \R_{>0}$ and $G = (\R_{>0}, \times)$ with group action $g\cdot \theta = g\theta$. Log-normal distributions with a fixed shape parameter, Rayleigh distributions, and exponential distributions can all be realized as~\cref{eq:Qcaldefn} with this group action. (bottom-left) $\Theta = \R$ and $G = \Aff_1^+(\R) = \{(A, b): A >0, b\in \R\}$ the affine group in one dimension, with the group action $(A,b) \cdot \theta = A\theta + b$. If $q_0$ is the unit normal distribution then $q^{(A,b)} = \mathcal{N}(b, A^2)$. (right) $\Theta = S^2$ and the special orthogonal group $G = \operatorname{SO}(3, \R)$ acts on the sphere by rotations. von Mises--Fisher distributions with a fixed concentration parameter can be constructed as a single orbit under the pushforward action of this group. } \label{fig:groupQexamples}
\end{figure}

The differential at a given $X\in \mathfrak{g}$ can be explicitly calculated by perturbing $g$ to $g\exp(tX)$ to get the real values $
\dd(\mathcal{E} \circ \varphi \circ L_g)\big|_{e}[X] = \frac{\dd}{\dd t}\mathcal{E}\left(q^{g \exp(tX)}\right)\Big|_{t = 0}.$
This is the directional derivative of the objective in~\Cref{eq:BLP} in the direction of $X$, (equivalently in the direction $\dd L_g\big|_e[X] \in T_gG$ at $g$).
For our variational objective, it is calculated as, 
\begin{equation}\label{eq:differentialObjective}
	\frac{\dd}{\dd t}\mathcal{E}(q^{g\exp(tX)}) \bigg|_{t =0}  = \EE_{\ttheta \sim q^g}[\nabla \ell(\ttheta)^\top (\operatorname{Ad}_g(X) \cdot \ttheta)] + \tau \xi(X).
\end{equation}
We include this short calculation in~\Cref{app:subsec:differential} for completeness, which is a straightforward application of the change of variables followed by differentiating under the integral, afforded to us by our Lie group setting. The term $\xi(X)$---which we introduced in the previous subsection---arises as the differential of the negative entropy term. The adjoint operator $\Ad_g: \mathfrak{g} \to \mathfrak{g}$ is the infinitesimal effect of conjugation as defined above, and can be concretely identified for each choice of Lie group $G$. The significance of \cref{eq:differentialObjective} is that it expresses the differential of the expected loss $\EE_g[\ell]$ with an expectation involving $\nabla \ell$. Thus, a perturbation in the distribution (a perturbation in $g$) is expressed as a perturbation in the parameters $\ttheta$. 

In summary, we update the group element using the group multiplication and the exponential map, in the direction of the steepest slope. In LG-BLR the bilinear form $B(\cdot ,\cdot )$ on $\mathfrak{g}$ determining this steepest slope is chosen as the pullback of the Fisher information metric on $\Qcal$. This pullback is well defined, and is independent of the group element $g$. As long as the Fisher metric is nonsingular~\cref{eq:FastestAscentOptimization} can be solved. It is possible that there are multiple solutions $\hat{Y}$, but we can always choose one among them. More details on obtaining the update direction are in~\Cref{app:subsec:gradient}. 
Furthermore we should point out that, the formalism of LG-BLR allows us to also simply start with a bilinear form $B$ on $\mathfrak{g}$ and that determines a left invariant metric on $G$, even if it doesn't necessarily come from the Fisher metric on $\Qcal$.

\subsection{On Applying Gradient Accumulation}

The Lie group update rule is an optimization on a manifold, nonetheless we can accumulate past gradients as mentioned in Section~3.5 of \cite{KiKh23}. This is thanks to the homogeneity of the Lie group, since we could formulate the updates as multiplications by ``small'' (i.e. near identity) group elements given by directions at $e \in G$. In other words, $\hat{Y}_t \in \mathfrak{g} = T_eG$ for all time steps $t$, regardless of the current position $g_t \in G$. So all the update vectors lie in the same vector space $\mathfrak{g}$. 

This generalizes momentum in the standard gradient descent setup on the flat parameter space $\R^P$. There as well, the gradient at time $t$ lies in the corresponding tangent space, $(\nabla_{\ttheta} \ell ) (\ttheta_t) \in T_{\ttheta_t}\R^P$, and an identification of all these tangent spaces is required for meaningfully operating with gradient vectors at multiple timepoints. The additive vector structure of $\R^P$ (as a special case of the group structure mentioned above) canonically identifies all these vector spaces with $\R^P$. For a deeper discussion regarding the mathematical underpinnings of what makes momentum possible with manifold learning, see~\Cref{app:connection}.

Furthermore, optimizers such as RMSprop, Adam~\citep{kingma2014adam},  or Lion~\citep{chen2023symbolic} not only accumulate past gradients, but also operate on the gradient vectors such as taking componentwise squares, adaptive componentwise scaling or taking signs. We are equally justified in doing such operations with LG-BLR in the Lie algebra $\mathfrak{g}$. The exponential map $\exp_G$ will still be able to retract the resulting vector onto the group. In \cite{nishidalog} this idea was successfully applied to the multiplicative group, where a Lion-like update was applied. Roughly speaking, the update used in that paper was of the form $g_{t + 1} \leftarrow g_t \exp(-\alpha \sign(M_t))$ where $M_t \in \mathfrak{g}$ is the standard momentum accumulation. This greatly increased the stability of the multiplicative update rule. 

In \cref{subsec:finalmomrem} we express this Lie group momentum in terms of natural gradients in BLR, in case the candidate posteriors are also an exponential family.

\subsection{An Example: Affine Groups and Gaussians} \label{sec:affex}

For each Lie group, LG-BLR corresponds to a different update. For example let $G = \Aff_P^+(\R)$ acting on $\Theta = \R^P$. Elements of $G$ are of the form $g = (A, \mathbf{b})$ where $A$ is an invertible $P\times P$ matrix with positive determinant and $\mathbf{b}$ is a vector in $\R^P$. The action is given by $(A,\mathbf{b}) \cdot \ttheta = A\ttheta + \mathbf{b}$. If $q_0 = \mathcal{N}(\ttheta \mid \vec{0}, I)$ then, the orbit of $q_0$ under the affine group action is the statistical manifold of all multivariate Gaussian distributions. Explicitly, $q^{(A,\vec{b})}(\ttheta) = \mathcal{N}(\ttheta | \vec{b}, AA^\top)$. The covariance matrix is $\Sigma = AA^\top$, and the mean is $\vec{m} = \vec{b}$. 
%The group multiplication is $(A_1, \mathbf{b}_1)(A_2, \mathbf{b}_2) = (A_1A_2, \mathbf{b}_1 + A_1\mathbf{b}_2)$ and this group operation satisfies $(g_1 g_2) \cdot \ttheta = g_1 \cdot (g_2 \cdot \ttheta)$. 

The LG-BLR updates the group elements $g$, along a tangent direction in $\mathfrak{g}$, which in turn gives an update in the Gaussian distribution $q^g \in \Qcal$. Elements of the affine Lie algebra (denoted by $\mathfrak{aff}_P$) can be written as $(X, \mathbf{x})$ where $X$ is a (not necessarily invertible) $P\times P$ matrix and $\mathbf{x} \in \R^P$. The bilinear form on this tangent space obtained as the pullback of the Fisher Information metric on $\Qcal$ via $\dd (\varphi \circ L_g)\big|_e$ is independent of $g$, and is given by
\begin{equation} \label{eq:LieAlgebraFisherAffine}
	\langle (X, \vec{x}), (Y, \vec{y}) \rangle  = 2\langle \Sym(X), \Sym(Y) \rangle_{\text{Frob}} + \vx^\top \vy.
\end{equation}
where $\Sym(X) = (X + X^\top)/2$, and the Frobenius inner product $\langle A, B \rangle_{\text{Frob}} = \Tr(A^\top B)$. This symmetrization exists because the unit normal distribution $q_0$ is invariant under rotations, and so $q^{gQ} = q^g$ for any orthogonal matrix $Q \in \operatorname{O}(P,\R)$. Therefore the bilinear form is degenerate in the direction of antisymmetric matrices.

The group can be represented as $(P + 1) \times (P+1)$ matrices $\left[\begin{smallmatrix}A&\mathbf{b} \\ \mathbf{0}^\top & 1 \end{smallmatrix}\right]$ and then the tangent matrices at identity are of the form $\left[\begin{smallmatrix} X & \mathbf{x} \\ \mathbf{0}^\top & 0 \end{smallmatrix}\right]$. The exponential map is given as
\begin{equation}\label{eq:affex}
	\exp( (X, \mathbf{x})) = \sum_{n = 0}^\infty \frac{1}{n!} \left[\begin{matrix}X&\mathbf{x} \\ \mathbf{0}^\top & 0 \end{matrix}\right]^n = I +  \sum_{n = 1}^\infty \frac{1}{n!} \left[\begin{matrix}X^n& X^{n-1} \mathbf{x} \\ \mathbf{0}^\top & 0 \end{matrix}\right] = \left(e^X, \frac{e^{X} - I}{X} \mathbf{x}\right).
\end{equation}

Now we specialize the differential of the objective functional in~\cref{eq:differentialObjective} to the case of the Affine group and Gaussians.
The adjoint map for matrix groups can be calculated by conjugation
\[
	\left[\begin{matrix} A & \vec{b} \\ \vec{0}^\top & 1 \end{matrix}\right]
	\left[\begin{matrix} X & \vec{x} \\ \vec{0}^\top & 0 \end{matrix}\right]
	\left[\begin{matrix} A\inv & - A\inv \vec{b} \\ \vec{0}^\top & 1 \end{matrix}\right]
	= \left[\begin{matrix}AXA\inv & A\vec{x} - AXA\inv \vec{b}\\ \vec{0}^\top & 0 \end{matrix}\right]
\]
and thus for a $\ttheta$ sampled from $q^{(A,\vec{b})}$, or in other words for $\ttheta = A\ttheta_0 + \vec{b}$ with $\ttheta_0 \sim q_0$, 
$$\operatorname{Ad}_{(A, \vec{b})}(X, \vx) \cdot \ttheta = AXA\inv (\ttheta - \vec{b})  + A\vec{x} = AX \ttheta_0 + A\vec{x}.$$ Therefore the differential of $\mathcal{E}$ is given by
\begin{align*}
	\dd (\mathcal{E} \circ \varphi\circ L_g)\big|_{(I, \vec{0})} [(X,\vec{x})]&= \int_{\Theta} (\nabla \ell(A\ttheta + \vec{b}))^\top (AX \ttheta + A\vec{x}) q_0(\ttheta)\dd\nu(\ttheta) - \tau \operatorname{Tr}(X)\\
	&= 2\Tr( \Sym(Y)^\top \Sym(X)) + \vy^\top \vx
\end{align*}
where 
\begin{equation}\label{eq:gaussian_y_update}
	\vy = A^\top \EE_{\ttheta \sim q_0}[ \nabla \ell(A\ttheta + \vec{b}) ] ,
\end{equation}
and
\begin{equation}\label{eq:gaussian_Y_update}
	Y = \tfrac{1}{2} A^\top \EE_{\ttheta \sim q_0} [\nabla \ell( A\ttheta + \vec{b}) \ttheta^\top ]  - \tfrac{\tau}{2} I .
\end{equation}

The inner product in~\cref{eq:LieAlgebraFisherAffine} on $\mathfrak{aff}_P$ induced via~\cref{eq:FisherBilinearOnGroup} from the Fisher information metric on $\mathcal{Q}$ gives this particular $(Y, \vec{y}) = (\hat{Y}, \hat{\vec{y}})$ as the direction that solves~\cref{eq:FastestAscentOptimization}.

In practice, we estimate the expectations using a finite number of MC-samples
\[
	Y \approx  \frac{1}{2K} \sum_{k = 1}^K A^\top \nabla \ell(A\ttheta_k + \vec{b}) \ttheta_k^\top - \tfrac{\tau}{2} I, \quad \vec{y} \approx \frac{1}{K} \sum_{k = 1}^K A^\top \nabla\ell(A\ttheta_k + \vec{b}), \text{ where } \ttheta_k \sim \mathcal{N}(\vec{0}, I)
\]
and update the current group element $g = (A,b)$ with these approximations as
\[
	A^\text{updated} = Ae^{-\alpha Y} , \qquad \vec{b}^{\text{updated}} = \vec{b} - A \frac{e^{- \alpha Y} - I}{-\alpha Y} \alpha \vec{y}. 
\]

Let us try to understand these updates intuitively. Up to a first order approximation in $\alpha$ the Gaussian's mean vector $\vec{b}$ is updated as 
$$\vec{b}^\text{updated} \approx \vec{b} - \alpha A \vec{y} = \vec{b} - \alpha A A^\top \EE_{q^{(A,\vec{b})}} [\nabla \ell] = \vec{b} - \alpha \Sigma  \EE_{q^{(A,\vec{b})}} [\nabla \ell].$$ So instead of the (average) gradient the update vector is modified by $\Sigma$, which has the effect of taking longer steps in directions where there is more uncertainty, and smaller steps where the Gaussian has less uncertainty. 
\begin{figure}
\centering
\includegraphics[width=0.5\textwidth]{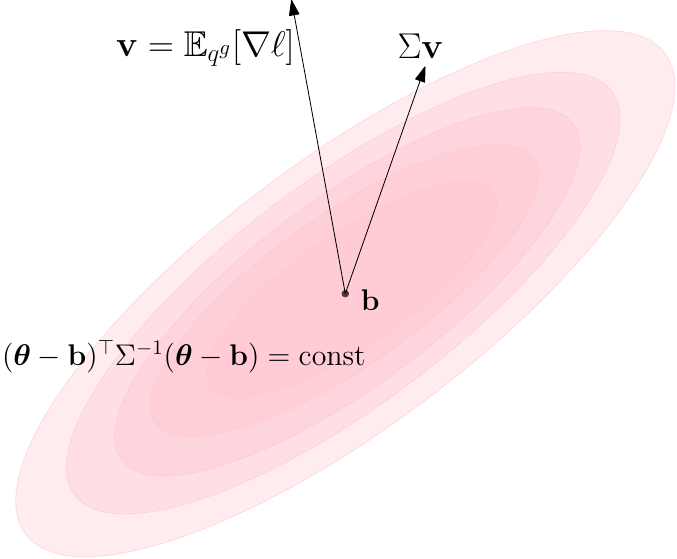}
\caption{The update vector is automatically modified so that the mean of the distribution takes larger steps in the direction of the uncertainty of the Gaussian, and smaller steps in direction of higher precision.} \label{fig:gaussianUpdate}
\end{figure} For the $A$ update, using integration by parts we rewrite 
\[
Y = \tfrac{1}{2}A^\top \EE_{q^{(A, \vec{b})}}[\nabla^2\ell] A  - \tfrac{\tau}{2} I.
\] 
This is not necessarily useful practically, as the computation of the Hessian is expensive, and instead we use the form in~\cref{eq:gaussian_Y_update}. However for conceptual analysis this form is useful. We see that the precision matrix of the Gaussian is iteratively replaced by the expected Hessian. Indeed, the precision matrix $\mathbf{S} = \Sigma\inv = A^{-\top} A\inv$, is updated as
\[
	\mathbf{S}^{\text{updated}} = A^{-\top}e^{\alpha (Y + Y^\top)} A\inv = A^{-\top} e^{\alpha (A^\top \EE_{q^{g}}[\nabla^2\ell] A - \tau I)} A\inv.
\]
The linear approximation to the exponential yields
\[
\mathbf{S}^{\text{updated}} \approx (1 - \alpha \tau ) \mathbf{S} + \alpha \EE_{q^{g}}[\nabla^2\ell],
\]
which is identical with the precision update in \citet[Eq.~12]{KhRu23}.
An advantage of the Lie group parametrization in terms of $A$ and $\vec{b}$ over the BLR is that it does not require any matrix inversions. Since $\vS$ is related to the Hessian matrix, this enables a matrix-inverse-free second-order optimization. Such algorithms may be desirable for large-scale deep learning~\citep{lin2023simplifying}, and our Lie group framework allows a natural way to derive these methods and also endow them with momentum. 

We also note that, at this stage we get the second order modification to BLR's precision matrix update---developed by~\citet{LiKh20}---for free by expanding the exponential to the second order. Indeed (using the shorthand $\mathbf{E} := \EE_{q^g}[\nabla^2\ell]$) we see
\begin{align*}
	&\mathbf{S}^{\text{updated}} = A^{-\top}e^{\alpha (Y + Y^\top)} A\inv = A^{-\top} e^{\alpha (A^\top \mathbf{E} A - \tau I)} A\inv \\
	&\approx A^{-\top} \left(I + \alpha  (A^\top \mathbf{E} A - \tau I)  + \tfrac{\alpha^2}{2} \left(A^\top \mathbf{E} \Sigma \mathbf{E} A - 2\tau A^\top \mathbf{E} A + \tau^2 I\right) \right)A^{-1}\\
	&= (1 - \alpha\tau) \mathbf{S} + \alpha \mathbf{E} + \tfrac{\alpha^2}{2} \left(\mathbf{E} \Sigma \mathbf{E} - 2\tau \mathbf{E} + \tau^2 \mathbf{S}\right) 
\end{align*}
which is identical (with $\tau = 1$) to the precision update $\mathbf{S} \leftarrow (1-\alpha) \mathbf{S} + \alpha \mathbf{E} + \frac{\alpha^2}{2} \hat{\mathbf{G}} \Sigma \hat{\mathbf{G}}$ proposed in the citation above, where $\hat{\mathbf{G}}= \mathbf{S} - \mathbf{E}$. The authors proposed this modification in order to preserve the positive definiteness constraint of the precision (equivalently the covariance) matrices throughout the optimization. The positivity follows since the second order truncation of the exponential is always positive definite, ensuring that the updates remain in $\Omega$, which is not guaranteed by the linear approximation. 

In the following sections, we will show that we recover the BLR in general, that is, that when we use the linear approximation for the exponential used in LG-BLR for candidate posterior families $\Qcal$ which are also open exponential families, then we get the BLR update. Our previous work~\citet{KiKh23} only proves this in particular cases, such as the Rayleigh and diagonal Gaussian family.

\section{Lie Group Structure of Exponential Families}

%Some distributional families are close-knit, it is possible to bridge their distributions by a reparametrization. For example sampling from the unit normal distribuiton $\varepsilon \sim \mathcal{N}(0,1)$\footnote{I don't think $\varepsilon$ is good to describe a general element of $\R$}, and applying $\varepsilon \mapsto \sigma \varepsilon + m$ gives us samples from $\mathcal{N}(m, \sigma^2)$. This means that the Gaussians are closed under the pushforward action, coming from the map $\varepsilon \mapsto  (m, \sigma)\cdot \varepsilon = \sigma \varepsilon + m$. More generally any multidimensional Gaussian can be obtained from the unit Gaussian by first applying an invertible linear map and then translating. In this case the action of the affine group on the samples induces a transitive action on the distributions. 

%The so-called reparametrization trick makes use of this structure to express gradients of expectations of functions (w.r.t. a variation of the distribution) in terms of expectation of the gradients of the functions (w.r.t. a variation in the sample space). Therefore it is interesting to characterize which common families of distributions are also given as translates of one of its members.

In~\Cref{sec:affex} above we saw that the Gaussians conform to the Lie group setup. They are also an exponential family of distributions. In general, the pushforward of a distribution $q_\llambda$ from an exponential family does not always correspond to another member $q_{\llambda'}$ for some other natural parameter $\llambda'$. In \emph{harmonic exponential families}---which we define below---we can always find such a $\llambda'$ explicitly. We are interested in open exponential families that have a transitive Lie group action given by pushforwards as above, that is, when the BLR and the LG-BLR settings coincide. The work of~\citet{ToYo21} allows us to characterize all such exponential families using harmonic exponential families. For the precise statement see~\Cref{thm:corollary_of_Tojo-Yoshino}. Using this concrete characterization we can express the Lie group updates in terms of updates in the natural parameters $\llambda$, with which we prove the connection between LG-BLR and BLR.

\subsection{Harmonic Exponential Families}

\begin{setup}[Transitive group action on parameter manifold]\label{set:GroupTransitive}
Assume $\Theta$ is the measurable parameter manifold with a choice of a nonzero base measure $\nu$. Assume a Lie group $G$ acts transitively on $\Theta$, and $\nu$ is relatively invariant with respect to this action with multiplier $\chi: G \to \R_{>0}$.  Because $G$ acts transitively, given any $\ttheta_0 \in \Theta$ we can identify $\Theta \cong G/H$ as spaces with a $G$-action, where $H = \Stab_G(\ttheta_0)$.  
\end{setup}
The last sentence of the setup says that we can associate any point $\ttheta \in \Theta$ with a group element $g \in G$ that satisfies $\ttheta = g\cdot \ttheta_0$.  The associated group element can be replaced with any other coset representative $gh \in gH\in G/H$.

What allows $(q_\llambda)^g$ to be another $q_{\llambda'}$ is a certain intertwining property of the sufficient statistic $\mathbf{T}: \Theta \to \primsp$ with respect to the action of $g$. On the target $\primsp$, group elements act on vector spaces by representations $\pi : G\to \GL(\primsp)$. So we seek a special kind of sufficient statistic such that $\mathbf{T}(g\cdot \ttheta)$ can be related to $\pi(g) \mathbf{T}(\ttheta)$.

The \emph{harmonic exponential family} construction yields this kind of an intertwining map. Start with a group representation $\pi$, meaning $\pi(g)$ are linear transformations on $\primsp$, i.e. matrices. Let $\aalpha : G \to \primsp$ be a $1$-cocycle of $\pi$, meaning that
\begin{equation}\label{eq:cocycleCondition}
	\aalpha(gh) = \pi(g) \aalpha(h) + \aalpha(g)
\end{equation}
for all $g,h \in G$. Furthermore, if we assume $\aalpha\big|_H \equiv 0$ then $\aalpha$ induces a map $\mathbf{T} : \Theta \to \primsp$, which also satisfies $\mathbf{T}(g\cdot \ttheta) = \pi(g) \mathbf{T}(\ttheta) + \text{const}$ (with the constant being precisely $\aalpha(g)$). With such a sufficient statistic $\mathbf{T}$, elements of the exponential family $q_\llambda \in \EF(\mathbf{T})$ satisfy
\begin{align}
	q_\llambda(g\inv \cdot \ttheta) &\propto \exp\left(-\langle \llambda, \mathbf{T}(g\inv \cdot \ttheta) \rangle\right) \notag \\
	&\propto \exp\left(-\langle \llambda, \pi(g)\inv \mathbf{T}(\ttheta) \rangle \right) \notag\\
	&= \exp\left( -\langle \pi^*(g) \llambda, \mathbf{T}(\ttheta) \rangle \right) \propto q_{\pi^*(g) \llambda}(\ttheta). \label{eq:harmonicPushforward}
\end{align}
Here $\pi^* : G \to \GL(\dualsp)$ is the dual representation defined via $\langle \pi^*(g) \llambda , \mathbf{v} \rangle = \langle \llambda , \pi(g)\inv \mathbf{v} \rangle$. This makes the exponential family closed under the pushforward action. And the new natural parameter is given explicitly by the multiplication of the matrix $\pi^*(g)$ with the old natural parameter vector $\llambda$. We collect the notation in the setup below.

\begin{setup}[Group representation and cocycles]\label{set:grpReprCocycles}
  Let $G$ be a connected Lie group and $H$ a closed subgroup, $\primsp$ a finite dimensional real vector space, and $\pi: G \to \GL(\primsp)$ a group representation. Let $\aalpha$ be a $1$-cocycle of $\pi$ (i.e. satisfying~\cref{eq:cocycleCondition}) vanishing on $H$. Using the notation from Setup~\ref{set:GroupTransitive}, $\aalpha : G \to \primsp$ induces a map $\mathbf{T}:\Theta \to \primsp$ using the identification of $\Theta \cong G/H$ as above. 
	
  We denote by $W_0 = W_0(G,H)$ the continuous group homomorphisms $\psi :G \to \R$ that vanish on $H$, that is $\psi(g_1g_2) = \psi(g_1) + \psi(g_2)$ and $\psi(h) = 0$ for all $h \in H$. These morphisms form a vector space. Furthermore they can be considered as real valued functions on $\Theta$, satisfying $\psi(g\ttheta) = \psi(g) + \psi(\ttheta)$.
\end{setup}

\begin{definition} 
Assume the setup and the notation from setups \ref{set:GroupTransitive} and \ref{set:grpReprCocycles}. A \emph{harmonic exponential family} on $\Theta$ is a family of distributions of the form
\begin{equation}\label{eq:harmonicExpDefn}
	q_{\eeta}(\ttheta) \propto \exp\left(-\langle \llambda, \mathbf{T}(\ttheta) \rangle - \psi(\ttheta)  \right) 
\end{equation}
parametrized by $\eeta = (\llambda, \psi) \in \dualsp \oplus W_0$.
\end{definition}

The normalizable distributions among~\cref{eq:harmonicExpDefn} have parameters $\eeta = (\llambda, \psi)$ satisfying $\llambda \in \Omega_{ e^{-\psi}\nu, \mathbf{T}} \subseteq \dualsp.$ Under the group pushforward action $\psi$ stays fixed; if $q_{\eeta'} = (q_\eeta)^g$ then $\eeta' = (\pi^*(g) \llambda, \psi)$. We will consider exponential families which are transitive under the group action (so that the LG-BLR updates can reach every member of the family). Thus,  the homomorphism component of the parameter is fixed in our posterior candidates, and we simply write $q_\llambda$  in place of $q_\eeta$, dropping $\psi$ from the notation. Note also that, if desired, the homomorphisms $\psi :G \to \R$ can be subsumed in the base measure, i.e. $\nu' = e^{-\psi} \nu$. And $\nu'$ is still a relatively invariant measure with respect to the action of $G$, with multiplier $\chi' = e^{-\psi}\chi$. Thus by only a change of base measure we assume from now on that $\psi \equiv 0$. 

\subsection{Some Examples of Harmonic Exponential Families}

For concreteness, we note how some well known exponential families are indeed realized as harmonic exponential families. The reader may skip these examples on a first reading.

\begin{example}[Gamma Distributions]
Gamma distributions are of the form
\[
	f_{\alpha, \lambda}(x) \dd x \propto x^{\alpha - 1} e^{-\lambda x}\dd x = \exp( - \lambda x + \alpha \log(x)) \tfrac{\dd x}{x}
\]
which is in the form of a harmonic exponential family with the invariant base measure $\dd\nu(x) = \dd x/x$, the homomorphism $\psi(x) = - \alpha \log(x)$, and sufficient statistic $T(x) = x$. Here the group is $G = \R_{>0}$ under multiplication and $H = \{1\}$ is the trivial subgroup. The representation is one dimensional with $\pi(g) = g$, and the relevant $1$-cocycle we use is $\aalpha(g) = g-1$. It satisfies $\aalpha(gh) = gh-1 = g(h-1) + (g-1) =  \pi(g)\aalpha(h) + \aalpha(g)$.  
\end{example}
Note that the group action fixes the shape parameter $\alpha$, i.e. $(f_{\alpha, \lambda})^g(x) = \tfrac{1}{g} f_{\alpha,\lambda}(\frac{x}{g}) = f_{\alpha, \frac{\lambda}{g}}(x)$. This amounts to the group action fixing the $\psi \in W_0$ component of the harmonic exponential family parametrization.

\begin{example}[Rayleigh Distribution]
  Let $G = \R_{>0}$, $H = \{1\}$ so that $\Theta \cong G/H \cong \R_{>0}$ as well. Consider the $1$ dimensional representation $\pi(g) = g^2$. The $1$-cocycle $\aalpha(g) = (g^2-1)/2$ associated to $\pi$ vanishes on $H$. The measure $\nu = x\dd x$ is relatively invariant with multiplier $\chi(g) = g^2$. (Alternatively we could have fixed the homomorphism $\psi(x) = -\log x$ and chosen $\nu = \dd x$.)  With these choices, the resulting harmonic exponential family is
	\[
		q_\lambda(x)\dd\nu(x) = \lambda x e^{-\lambda \frac{x^2}{2}} \dd x.
	\]
	The normalizable set of natural parameters are $\Omega_{\nu, \aalpha} = \{\lambda \in \R : \lambda > 0\}$. The family is closed under the pushforward action by $g$, as $$g_\sharp(q_\lambda\dd \nu) = \frac{1}{\chi(g)}q_\lambda(g\inv x) \dd\nu(x) = \frac{x}{g^2}\lambda e^{- \lambda x^2/2g^2}\dd x = q_{\frac{\lambda}{g^2}}(x) \dd\nu = q_{\pi^*(g)\lambda}(x) \dd\nu.$$ Here the dual action on $\dualsp\cong \R$ is given by the $1\times 1$ matrix $\pi^*(g) = \pi(g\inv)^\top =  1/g^2.$ 
\end{example}

\begin{example}[Poincar\'e distribution]
  The following family is called the family of Poincar\'e distributions on the upper half plane $\mathcal{H}=\{z=x+iy \mid x, y \in \R, y>0\}$. 
  \begin{align*}
    \left\{\frac{De^{2D}}{\pi} \exp\left( - \frac{a(x^2+y^2)+2bx+c}{y}\right)\frac{dxdy}{y^2}\right\}_{\left[\begin{smallmatrix}a & b \\ b & c\end{smallmatrix}\right]\in \Sym^+ (2,\R) }, 
  \end{align*}
  where $D=\sqrt{\det \left[\begin{smallmatrix}a & b \\ b & c\end{smallmatrix}\right]}=\sqrt{ac-b^2}$ and $\Sym^+(2,\R)$ denotes the set of positive definite symmetric matrices of size 2.  
  This family is a harmonic exponential family on $\mathcal{H}$. 
  Let $G=\SL_2(\R)$ and $H=\SO_2(\R)$. 
  The Lie group $G$ acts transitively on the upper half plane $\mathcal{H}:=\{z=x+iy \mid x,y\in \R, y>0\}$ by linear fractional transformation. 
  Then we have $\Stab_G(i)=H$. Therefore we can identify $\SL_2(\R)/\SO_2(\R)$ with $\mathcal{H}$ via the following $G$-equivariant map: 
  \begin{align*}
    \mathcal{H}& \to \SL_2(\R)/\SO_2(\R),\\ z=x+iy &\mapsto \begin{bmatrix}\sqrt{y} & \frac{x}{\sqrt{y}}\\ 0 & \frac{1}{\sqrt{y}} \end{bmatrix} 
  \end{align*}
  Let $\pi\colon G\to \GL(\Sym(2,\R))$ (so $\primsp = \Sym(2, \R)$ is the three dimensional vector space of matrices) be the representation defined by 
  $\pi(g)S= g S g^T$ for $S\in \Sym(2,\R)$, and $\aalpha\colon G\to \Sym(2,\R)$ the 1-cocycle given by $\aalpha(g):=\pi(g)I_2-I_2$, which satisfies $\aalpha|_H=0$. 
  We take an invariant measure $\frac{dx dy}{y^2}$ with respect to the $\SL_2(\R)$-action, and $\psi \equiv 0$ (The homomorphism from $\SL_2(\R)$ to $\R$ must be zero). In this case $\dualsp$ can also be identified with $\primsp$ and normalizability gives us positive definite symmetric matrices as $\Sym^+(2, \R) = \Omega \subset \dualsp$. 
  
The resulting harmonic exponential family is the family of Poincar\'e distributions.  
  By the change of variable, 
  \begin{align*}
    \Sym^+(2,\R)&\simeq \R_{>0} \times \mathcal{H}, \\
    \begin{pmatrix}a&b\\b &c \end{pmatrix}&\mapsto (\rho,\xi+i\eta):=\left(\sqrt{ac-b^2},  -\frac{b}{a}+ i \frac{\sqrt{ac-b^2}}{a}\right)
  \end{align*}
the Poincar\'e distribution can be written as the following ``normal-like'' form
\begin{align*}
  \frac{\rho}{\pi}\exp\left( -\frac{\rho}{y\eta}|z-w|^2 \right) \frac{dxdy}{y^2}, 
\end{align*}
where $\rho \in \R_{>0}$, $w=\xi+i\eta\in \mathcal{H}$. 
See  \cite{NielsenOkamura2024}, \cite{TojoYoshino_Poincare2025} for more information. 
\end{example}

\begin{example}
	Multivariate Gaussian distributions form a harmonic exponential family. In this case $G = \Aff_n^+(\R)$, realized as square matrices of size $(n + 1)$ of the form $g = \left(\begin{smallmatrix}A&  \vec{b} \\ \vec{0}^\top & 1 \end{smallmatrix}\right)$, for $A \in \GL_n(\R)$ with positive determinant and $\vec{b} \in \R^n$. The subgroup $H \cong \GL_n^+(\R)$ is embedded as $\left\{\left(\begin{smallmatrix} A & \vec{0} \\ \vec{0}^\top & 1 \end{smallmatrix}\right): A\in \GL_n^+(\R) \right\}.$ The sample space is $\Theta := G/H \cong \R^n$, where from the quotient we can pick representatives with $A = I$ and identify any $\vec{b} \in \R^n$ with $\left\{\left(\begin{smallmatrix} I & \vec{b} \\ \vec{0}^\top & 1 \end{smallmatrix}\right) : \vec{b} \in \R^n \right\}$, and vice versa. Set $\psi \equiv 0$. 
	
	The action of $g \in G$ on $\ttheta \in \R^n$ is  $g \cdot \ttheta = A \ttheta + b$, which can also be realized as the matrix multiplication by viewing $\ttheta\in \R^n$ as an element of $G/H$. 
	
	Let $\primsp = \Sym(n + 1, \R)$, the vector space of $(n + 1)$-dimensional symmetric matrices. The (standard) Frobenius inner product on $\primsp$ is $\langle A, B  \rangle_{\text{Frob}} = \Tr(A^\top B) = \Tr(AB)$. This is the same as the Euclidean inner product on the coordinates of the matrices, when flattened. Under this inner product the dual space can be given by the same symmetric matrix space
\[
	\dualsp \cong  \Sym(n + 1, \R) =  \left\{\begin{pmatrix} \Lambda &\llambda\\ \llambda^\top  & c 
	\end{pmatrix}: c \in \R, \llambda \in \R^n, \Lambda \in \Sym(n, \R) \right\}.
\]

$G$ acts on $\primsp$ by transpose conjugation, i.e.\ for symmetric matrices $S\in \primsp$ we have the representation $\pi(g) S = g S g^\top$ of $\Aff_n^+(\R)$. The dual representation on $\dualsp$ is given by $\pi^*(g)S = g^{-\top} S g^{-1}$ for $S$ now viewed as an element of $\dualsp$.

A  $1$-cocycle for this representation is $\aalpha(g) := (g \vec{e})(g \vec{e})^\top - \vec{e} \vec{e}^\top$ for $\vec{e}$ equal to the $(n + 1)$\textsuperscript{th} standard basis vector, i.e.\ $\vec{e} = \vec{e}_{n + 1} = \left(\begin{smallmatrix} \vec{0}\\ 1 \end{smallmatrix} \right)$. 
We encourage the reader to check the cocycle condition $\aalpha(g\cdot  \ttheta) = \pi(g) \aalpha (\ttheta) + \aalpha(g)$ as well as that $\aalpha$ vanishes on $H$. The sufficient statistic coming from this $1$-cocycle is therefore
\[
	\vec{T}_\aalpha (\ttheta) = \begin{pmatrix} \ttheta \ttheta^\top &  \ttheta\\ \ttheta^\top & 0  \end{pmatrix}  = \begin{pmatrix} \ttheta \\ 1\end{pmatrix} \begin{pmatrix} \ttheta^\top &  1 \end{pmatrix} - \begin{pmatrix} 0 & 0 \\ 0 & 1 \end{pmatrix}.
\]

The distributions in the harmonic exponential family generated from this data are
\[
	q_{\left(\begin{smallmatrix}\Lambda & \llambda \\ \llambda^\top & c \end{smallmatrix}\right)}(\ttheta) \propto \exp\left(- \left\langle \begin{pmatrix}\Lambda & \llambda \\ \llambda^\top & c \end{pmatrix},  \begin{pmatrix} \ttheta \ttheta^\top &  \ttheta\\ \ttheta^\top & 0  \end{pmatrix} \right\rangle_{\text{Frob}}\right) = \exp\left(- \ttheta^\top \Lambda \ttheta  - 2\llambda^\top \ttheta \right).
\]
On the second equality we used the cyclic property of trace. The right hand side is normalizable if and only if $\Lambda$ is positive definite (defining $\Omega_{\dd\ttheta, \vec{T}}$). Furthermore this is precisely the multivariate Gaussian distribution $\mathcal{N}(\ttheta \mid \vec{m}, \Sigma)$ with $\Sigma = \tfrac{1}{2}\Lambda^{-1}$ and $\vec{m} = - \Lambda\inv\llambda $.

Note that $c$ does not appear on the right hand side, so we do not have a minimal representation, this can be fixed by considering the subspace of $\primsp$ with lower right corner equal to $0$. This subspace is preserved by $\pi(g)$ with $g \in \Aff_n^+(\R)$.

The dual representation $\pi^*(g)$ can also be explicitly calculated by matrix multiplication
\[
	\pi^*(g) \begin{pmatrix}
	\Lambda & \llambda \\ \llambda^\top & 0
	\end{pmatrix}
	= g^{-\top} \begin{pmatrix}
	\Lambda & \llambda \\ \llambda^\top & 0
	\end{pmatrix} g^{-1} 
	\!=\!  \begin{pmatrix}
	A^{-\top} \Lambda A^{-1} & -A^{-\top} \Lambda A^{-1} \vec{b} \!+\! A^{-\top} \llambda\\ 
	-\vec{b}^\top A^{-\top} \Lambda A^{-1} \!+\!  \llambda^\top A^{-1}  & 
	\end{pmatrix}
\]	
where we do not need to keep track of the lower right entry.
This is the action on the natural parameters, downstream of the affine action $\ttheta \mapsto A\ttheta + \vec{b}$ on the parameters. Particularly
\[
	\Lambda^{\text{new}} = A^{-\top} \Lambda A^{-1} \qquad \text{ and } \qquad \llambda^{\text{new}} = - \Lambda^{\text{new}} \vec{b} + A^{-\top} \llambda
\]	
from which we can write in the new Gaussian's mean-covariance coordinates $\Sigma^{\text{new}} = A\Sigma A^\top $ and $\vec{m}^{\text{new}} = A\vec{m} + \vec{b}$. This sanity check confirms the mapping of natural parameters.

\end{example}

%Exponential families constructed in this way are called \emph{harmonic exponential families}\footnote{the definition of harmonic exponential families is vague. For example, the information of the parameter space is excluded. Moreover, in Tojo-Yoshino construction consider all the nonzero relatively $G$-invariant measures so that the obtained family does not depend on the choice of a relatively $G$-invariant measure. However, in the explanation above, the constructed family depends on the choice of the base measure in general.}. 

\subsection{Characterizing Exponential Families Closed under a Group Action}

The work of \cite{ToYo21} characterizes the exponential families closed under the pushforward action of a Lie group as harmonic exponential families.

%\ckoichi{ The settings of Theorem~\ref{thm:corollary_of_Tojo-Yoshino}and Lemma~\ref{fact:Tojo-Yoshino_GSI25-paper} are the same, so the setting should be written not inside of Theorem~\ref{thm:corollary_of_Tojo-Yoshino} but before it, and be reused in Theorem~\ref{fact:Tojo-Yoshino_GSI25-paper} Setting: Let $G$ be a Lie group with finitely many connected components, $H$ a closed subgroup of $G$ and $\Theta:=G/H$ a homogeneous space of $G$.  We assume that $\Theta$ admits nonzero relatively $G$-invariant measures. }

\begin{theorem}[Tojo \& Yoshino]\label{thm:corollary_of_Tojo-Yoshino}
 	Assume Setup \ref{set:GroupTransitive} and its notation. Also assume that $\mathcal{Q}$ is an exponential family on $\Theta$ which is closed under the pushforward action of $G$, and the action is transitive on $\Qcal$. Then $\Qcal$ is a harmonic exponential family. 
 	In other words, if the harmonic exponential family is constructed from $\pi, \aalpha$ in the notation of Setup \ref{set:grpReprCocycles}, then $\mathcal{Q} = \{q_\llambda e^{-\psi}: \llambda \in  \Omega(\Qcal) \}$ where 
 	\begin{equation}\label{eq:orbitOmega}
 	\Omega(\mathcal{Q}) =\{\pi^*(g) \llambda_0 : g \in G\} \subseteq \Omega
 	\end{equation}
 	for some $\llambda_0 \in  \Omega = \Omega_{\nu e^{-\psi}, \mathbf{T}}$, and $\psi \in W_0$. Here $\mathbf{T}$ is obtained by factoring $\aalpha: G\to V$ through the canonical projection $G \twoheadrightarrow G/H \cong \Theta$. The homomorphism $\psi: G\to \R$ is fixed throughout and can be subsumed in the base measure $\nu$ if desired.
 \end{theorem}
 
Therefore we can concretely write the distributions in the intersection of the BLR setting of~\citet{KhRu23} and the LG-BLR setting of~\citet{KiKh23}.

\begin{proof}[Proof of Theorem~\ref{thm:corollary_of_Tojo-Yoshino}]
  Take any $q_0\in\mathcal{Q}$. 
%  Since the pushforward action of $G$ for distributions of the form \cref{eq:harmonicPushforward} is given by the dual representation on the parameter space, we need only show that one element of $\mathcal{Q}$ is of the form \eqref{eq:harmonicExpDefn}. 
  By Lemma~\ref{fact:Tojo-Yoshino_GSI25-paper}, there exist a relatively $G$-invariant measure $\nu$ on $\Theta$, a finite dimensional representation $\pi\colon G\to GL(\primsp)$ and a 1-cocycle $\aalpha$ of $\pi$ which vanishes on $H$ (inducing $\vec{T}: \Theta \to \primsp$) such that $q_0=q_{\llambda_0}e^{-\psi}$ for some $\llambda_0\in \Omega_{\nu,\vec{T}}$ and $\psi \in W_0$.  The orbit of $\llambda_0$ under $\pi^*(g)$ forms $\Omega(\mathcal{Q})$, and is contained in the full harmonic exponential family formed by $\pi$ and $\aalpha.$ The action of $g$ fixes $\psi$ and it can be subsumed in the measure $\nu$. 
 % In fact, by taking a representation $\pi$ and a 1-cocycle $\aalpha(g)=\pi(g)\vv_0-\vv_0$ as in Lemma~\ref{fact:Tojo-Yoshino_GSI25-paper} such that the obtained harmonic family includes $\mathcal{Q}$, 
 % we get $q_0=q_{\llambda_0}(\theta)$ for some $\llambda_0\in \Omega_{\nu,\aalpha}$ and a nonzero relatively $G$-invariant measure $\nu$ on $\Theta$. 
\end{proof}

\begin{lemma}[{\citet[Theorem~2]{ToYo25}}]\label{fact:Tojo-Yoshino_GSI25-paper}
%    Let $G$ be a Lie group with finitely many connected components, $H$ a closed subgroup of $G$ and $\Theta:=G/H$ a homogeneous space of $G$.  We assume that $\Theta$ admits nonzero relatively $G$-invariant measures. 
Assume the notation in Setup \ref{set:GroupTransitive}.
  Let $\mathcal{P}\subset \mathcal{R}(\Theta)$ be an exponential family on $\Theta$, closed under the pushforward action of $G$. Here $\mathcal{R}(\Theta)$ denotes the set of all Radon measures on $\Theta$. 
  
 % Then, there exists a finite dimensional real representation $\pi\colon G\to GL(\primsp)$ and an $H$-fixed cyclic vector $\vv_0\in \primsp^H$ such that the family obtained by applying $G/H$-method to $\pi$ and $\aalpha$ includes $\mathcal{P}$, where $\aalpha\colon G\to \primsp$ is given by $\aalpha(g)=\pi(g)\vv_0-\vv_0$. 
 Then $\mathcal{P}$ is contained in a harmonic exponential family, with notation as in Setup \ref{set:grpReprCocycles}. 
In particular, there exist $\pi\colon G\to \GL(\primsp)$ and 1-cocycle $\aalpha \colon G\to \primsp$ for $\pi$ such that any $q_0\in \mathcal{P}$ is of the form in~\cref{eq:harmonicExpDefn}. 
 %In particular any $q_0\in \mathcal{P}$ is of the from \eqref{eq:harmonicExpDefn} for some finite dimensional group representation $\pi : G \to \GL(\primsp)$, $\aalpha: G\to\primsp$ a 1-cocycle for $\pi$, and some  $\llambda_0 \in \dualsp$ and $\psi \in W_0$.
\end{lemma}

The heuristic proof for the above theorem is as follows. Members of an exponential family are of the form $q_\llambda \propto e^{-f_\llambda}$ where $f_\llambda(\ttheta) = \langle \llambda, \vT(\ttheta) \rangle$ is in a finite dimensional function space $\mathcal{F}$ spanned by $T_1(\ttheta), T_2(\ttheta), \ldots, T_d(\ttheta)$ and the constant function $1$. Therefore if an exponential family is closed under the pushforward action of $g \in G$, then space of functions $\mathcal{F}$ is closed under the action $(\pi(g) \cdot f)(\ttheta) := f(g\inv \cdot \ttheta)$. From this representation of $G$ on $\mathcal{F}$ we construct  and $\aalpha(g) = \pi(g) f_0 - f_0$ for some nonconstant $f_0 \in \mathcal{F}$. This is the heuristic behind why exponential families closed under the action of $G$ admit such an algebraic structure.

\subsection{Properties of Harmonic Exponential Families}

Given $\mathbf{T}$ as above---related to a base measure $\nu$, a representation $\pi$ and its $1$-cocycle $\aalpha$---start with a probability distribution $q_{\llambda_0} \in \EF_\nu(\mathbf{T})$ for $\llambda_0 \in \Omega$. Given this distribution as the starting point (denoted by $q_0$ in Section \ref{sec:LieGroupBLR} above), form its $G$-orbit $\Qcal$. Let $q_{\llambda'}(\ttheta) = (q_{\llambda_0})^g(\ttheta) = \frac{1}{\chi(g)}  q_{\llambda_0}(g\inv \cdot \ttheta)$, then direct computation yields,
\begin{align*}
\llambda' &= \pi^*(g) \llambda_0\\
\mmu' &= \pi(g)\mmu_0 + \aalpha(g)\\
A(\llambda') &= A(\llambda_0) + \langle \llambda_0, \aalpha(g\inv) \rangle +  \log \chi(g) 
\end{align*} 
where $\mmu', \mmu_0 \in \primsp$ are the expectation parameters of $q_{\llambda'}$ and $q_{\llambda_0}$. Note that the matrix multiplying the expectation parameter is dual (inverse transpose)  to the one multiplying the natural parameter, so the parameters $\llambda$ and $\mmu$ are not only duals of one another in Fenchel sense but also in how they get transformed  by a Lie group element.

\begin{figure}[h!]
\begin{center}
\includegraphics[width = 0.34\textwidth]{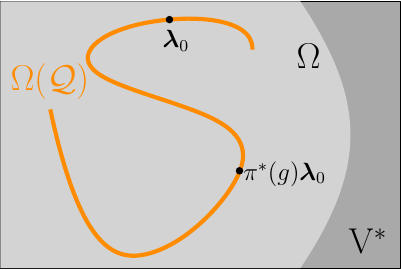}

\caption{The orbit $\Omega(\Qcal)$ of $\llambda_0\in \Omega$ under the action of $\pi^*$ stays in $\Omega$. %By running through various $\lambda_0\in \Omega$, the orbits disjointly cover $\Omega$. %Here the orbit of $\lambda_0$ is not technically $\Qcal$ but rather the set of natural parameters of distributions that are in $\Qcal$. There is a one to one correspondence between the two sets (we did assume the exponential family was minimal).
} \label{fig:Gorbit}
\end{center}
\end{figure}
The orbit $\Omega(\mathcal{Q}):= \{\pi^*(g) \llambda_0 : g\in G \} \subseteq \Omega$ of a given $\llambda_0 \in \Omega$ forms the candidate natural parameters to choose from. For BLR, since the update in~\cref{eq:BLR1} is linear in $\llambda$, in order to guarantee $\llambda^{\text{updated}} \in \Omega(\Qcal)$ at least for some small enough step size, we need the orbit to be open (i.e. for $\Qcal$ to be an open exponential family), or be relatively open in an affine subset of $\Omega$. For example, for normal distributions $\Omega = \left\{\llambda = (\lambda_1, \lambda_2) \middle| \lambda_2>0\right\}$ is open, and the latter happens for fixed variance Gaussian distributions. In the latter case, as we have shown in Example \ref{ex:1dgaussian}, we can switch to a minimal realization $R_{\text{min}}$ and get an open $\Omega_{R_{\text{min}}}(\Qcal)$. For LG-BLR, the openness of the parameter space is not required, since $\pi^*(g^{\text{updated}})\llambda_0$ remains on the orbit $\Omega(\Qcal)$, even for curved exponential families. We depicted one such illustrative natural parameter orbit above in~\Cref{fig:Gorbit} with the orange curve.

\begin{figure}[h!]
\begin{center}
\includegraphics[width = \textwidth]{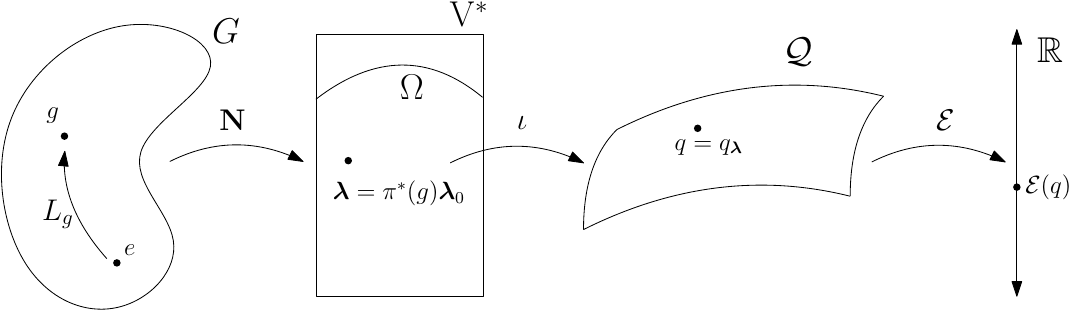}
\caption{Composition of maps that turn the variational inference problem into an optimization problem near the group identity element. We optimize $\mathcal{E}(q_{\pi^*(g) \llambda_0})$ by taking a small step $g \mapsto g \exp(-\alpha Y)$ for some $Y \in \mathfrak{g}$. We choose $Y = \hat{Y}$ as the direction of fastest ascent for $\mathcal{E}\circ (\iota \circ \mathbf{N}) \circ L_g$ at  $e$ computed with respect to a bilinear form on $\mathfrak{g} = T_eG$ which is the pullback of the Fisher information metric.} \label{fig:refinedGroupComposition}
\end{center}
\end{figure}

The harmonic exponential family framework gives us a refinement of Figure \ref{fig:groupComposition} whereby we factor the parametrization $\varphi: G \to \mathcal{Q}$ into $\varphi = \iota \circ  \mathbf{N}$. Here (having fixed a $\llambda_0$) we define the map $\mathbf{N} : G \to  \dualsp$ as \begin{equation}\label{eq:Ndefn}
\mathbf{N}(g) := \pi^*(g) \llambda_0.
\end{equation}
This gives the natural parameter associated to a group element, and
\begin{align*}
	\iota : \Omega_{\nu,\vT} &\longrightarrow \EF_\nu(\vT)\\
	\llambda &\longmapsto q_\llambda
\end{align*}
produces the distribution when given a natural parameter.

\subsection{Lie Group Update on Harmonic Exponential Families}
After establishing the explicit map between the natural parameters and the group elements, we can write the Lie group update in terms of $\llambda$. If we start at $\llambda = \mathbf{N}(g)$, applying $\mathbf{N}$ to both sides of the group update given by~\Cref{eq:gupdate}, we get the new natural parameters as
\begin{equation}\label{eq:theGroupUpdateInLambdas}
	\llambda^{\text{updated} } = \mathbf{N}(g \exp(-\alpha \hat{Y})),
\end{equation}
where $\hat{Y}$ is given as above in~\Cref{eq:FastestAscentOptimization}, and $\alpha >0$ is the step size. 

As the main result of this paper, we show that for an open exponential family $\Qcal$ that is also in the LG-BLR setting, the ``linearization'' of the exponential map in~\Cref{eq:theGroupUpdateInLambdas} gives the BLR update in $\llambda$, i.e.\ it gives the update in~\cref{eq:BLR1}. 

As an aside, in~\Cref{app:subsec:curvedlinearization} we see what happens in the case of linearizing~\Cref{eq:theGroupUpdateInLambdas} for curved exponential families. The LG-BLR setting does not require $\Qcal$ to be an open exponential family, it can be curved. In such a case, it turns out that the linearization of the $\llambda$ update is the orthogonal projection (with respect to the Fisher metric at $\llambda \in \Omega$) of the BLR update term onto the tangent space of $\Omega(\Qcal)$ embedded in $\Omega$.

In the next section we precisely write what we mean by the term linear approximation to the exponential in~\Cref{eq:theGroupUpdateInLambdas}.

\section{BLR as a Linear Approximation of LG-BLR}

\subsection{The Linearization} 

Assume the notation in Setup \ref{set:GroupTransitive}, and that there is an open exponential family $\Qcal$ that is closed and in fact a single orbit under the Lie group's pushforward action. In short we are in the joint setting where both BLR and LG-BLR are applicable. By~\Cref{thm:corollary_of_Tojo-Yoshino} this family is a subset of a harmonic exponential family. Write this harmonic exponential family using the notation of Setup \ref{set:grpReprCocycles}.

We now show that the linearization of \Cref{eq:theGroupUpdateInLambdas} gives the BLR given by~\Cref{eq:BLR1}. 

The representation $\pi^* : G \to \GL(\dualsp)$ is a homomorphism of Lie groups and its differential (at identity) is a linear map $\dd\pi^* : \mathfrak{g} \to \mathfrak{gl}(\dualsp)$. Here $\mathfrak{gl}(\dualsp)$ is simply the space of all linear maps from $\dualsp$ to itself, i.e.\ in coordinates, all $\dim(\dualsp)\times \dim(\dualsp)$ matrices. The exponential maps, the group homomorphism $\pi^*$ and its differential $\dd\pi^*$ intertwine as follows
\begin{equation}\label{eq:differentialpi}
	\pi^* (\exp( - \alpha Y)) = \exp\left(-\alpha ( \dd\pi^* [Y])\right).
\end{equation}
This is in fact true for all group homomorphisms, and not just $\pi^*$. 

The exponential map on the right is matrix exponentiation $\exp: \mathfrak{gl}(\dualsp) \to \GL(\dualsp)$. It can be given as a Taylor series, and its linear approximation is
\begin{equation}\label{eq:exponentialapprox}
	\exp(X) = e^{X} = \sum_{n = 0}^\infty  \frac{X^n}{n!} \approx I + X.
\end{equation}

Using this as our linearization 
%and recalling the definition \eqref{eq:Ndefn} of $\mathbf{N}$ as the parametrization $\llambda \in \dualsp$ by group elements, 
we have 
\begin{align} 
\llambda^{\text{updated}} &= 	\mathbf{N}\left(g \exp(-\alpha Y)\right) = \pi^*(g \exp(-\alpha Y)) \llambda_0 = \pi^*(g) \pi^*(\exp(-\alpha Y)) \llambda_0 \notag \\
	&= \pi^*(g) e^{-\alpha \dd\pi^*[Y]} \llambda_0 \approx \pi^*(g) \left(I - \alpha \dd\pi^*[Y] \right) \llambda_0.\label{eq:theLinearizationOfLambdaUpdate}
\end{align}
The first equality is from~\cref{eq:theGroupUpdateInLambdas}, and the second is the definition of $\mathbf{N}$ in~\cref{eq:Ndefn}. For the third equality we use that $\pi^*$ is a group homomorphism. The fourth equality is due to~\cref{eq:differentialpi}, and the final step is the approximation \cref{eq:exponentialapprox}.

For now, we kept $Y$ arbitrary. In the following subsections we show that if $Y = \hat{Y}$ is the direction of steepest ascent with respect to the Fisher metric, we recover the BLR in~\cref{eq:BLR1}.

\subsection{The Direction of Motion}\label{subsec:dirMotion}
Now let us connect $\hat{Y}$ with $\dd (\mathcal{E}\circ \iota)\big|_\llambda$.  Recall that $\varphi = \iota \circ \mathbf{N}$. We call $\dd (\mathbf{N} \circ L_g)\big|_e = R$ for ease of notation. $R: \mathfrak{g} \to T_\llambda \Omega \cong \dualsp$ is a linear map, and we have $\dd(\mathcal{E} \circ \iota\circ  \mathbf{N}\circ L_g) = \dd (\mathcal{E} \circ \iota)\big|_{\llambda} R$.
\begin{remark}
In~\cref{subsec:variationalLearning} above we made the identification $\mathcal{E}(\llambda) = \mathcal{E}(q_\llambda)$ which amounts to omitting $\iota$. Therefore $\dd\mathcal{E}\big|_{\llambda}$ in~\cref{eq:BLR1} corresponds to $\dd (\mathcal{E}\circ \iota)\big|_{\llambda}$ in our notation. 
\end{remark}

The linear map $R$ also relates the bilinear form $B$ on the Lie algebra $\mathfrak{g}$ with the Fisher metric in the natural parameter space: 
\[
	B(X,Y) = \omega_{\text{Fisher}}^\lambda (RX, RY) \text{ for all } X, Y \in \mathfrak{g}.
\]

\begin{proposition}\label{prop:FisherInverseGradient}
Assume that $\Qcal$ is an open exponential family and that the pushforward action of Lie group $G$ is transitive on $\Qcal$. Adopt the above notation. Then any $\hat{Y} \in \mathfrak{g} = T_eG$ solving $B(\hat{Y}, \cdot) = \dd(\mathcal{E} \circ \iota)\big|_{\llambda} R$ satisfies 
\[
	R\hat{Y} = \hat{\vv} = {F_\llambda}\inv \dd(\mathcal{E}\circ \iota)\big|_\llambda = \widetilde{\nabla}\mathcal{E}(q_\llambda)\big|_{\llambda}.
\]
\end{proposition}
\begin{proof}
It is possible, in general, that $R$ does not map onto all of $\dualsp$, but to a subspace $\operatorname{Im}(R) = W \subseteq \dualsp$. This is not the case here, as we assume an open exponential family. Put $R\hat{Y} = \hat{\vv}$, and also writing $RX = \vv$ (which runs over any $\vv \in \dualsp$ as $R$ is onto), we have $$\omega_{\text{Fisher}}^\llambda(\hat{\vv}, \vv) = \omega_{\text{Fisher}}^\llambda(R\hat{Y}, RX) = B(\hat{Y}, X) =  \dd(\mathcal{E} \circ \iota)\big|_\llambda [RX] = \dd(\mathcal{E} \circ \iota)\big|_\llambda [\vv].$$  
Since this should hold for every $\vv$ and since $\omega_{\text{Fisher}}^\llambda(\hat{\vv}, \vv) = \hat{\vv}^\top F_\llambda \vv$, we can see that $$F_\llambda \hat{\vv} = \dd(\mathcal{E} \circ \iota)\big|_\llambda$$ and this proves the result.
\end{proof}

\subsection{Putting it all Together} 
\label{subsec:wrapup}

Let us again write the map $R = \dd(\mathbf{N}\circ L_g)\big|_e$, and express $R\hat{Y}$ in terms of its $\pi^*$ dependence,
\begin{multline}
	R\hat{Y} = \dd(\mathbf{N} \circ L_g)\Big|_e[\hat{Y}] = \frac{\dd}{\dd t} \left( (\mathbf{N}\circ L_g) (\exp(t\hat{Y})) \right)\Big|_{t = 0} = \frac{\dd}{\dd t} \pi^*(g \exp(t\hat{Y}))\llambda_0 \Big|_{t=0} \\
	=\frac{\dd}{\dd t} \pi^*(g) \exp(t(\dd\pi^*[\hat{Y}])) \llambda_0 \Big|_{t = 0} = \pi^*(g) \dd\pi^*[\hat{Y}] \llambda_0 \label{eq:RYInTermsOfPiVee}
\end{multline}

Using~\Cref{eq:theGroupUpdateInLambdas,eq:theLinearizationOfLambdaUpdate} and the calculation in~\cref{eq:RYInTermsOfPiVee} above we express the linearization of the Lie group update rule as
\begin{align*}
	\llambda^{\text{updated}} = \mathbf{N}(g \exp(-\alpha \hat{Y})) &\approx \pi^*(g) (I - \alpha (\dd\pi^*[\hat{Y}])) \llambda_0 \\
	&= \pi^*(g) \llambda_0 - \alpha \pi^*(g) (\dd\pi^*[\hat{Y}]) \llambda_0 \\
	&= \llambda - \alpha R\hat{Y}\\
	&= \llambda -\alpha \widetilde{\nabla}_{\llambda} \mathcal{E}(q_\llambda).
\end{align*}
In the final line we used Proposition \ref{prop:FisherInverseGradient}.

This establishes the main theorem.
\begin{theorem}\label{thm:mainTheorem}
	Given a machine learning model with parameter space $\Theta$ and loss function $\ell$, assume that a Lie group $G$ acts transitively and smoothly on $\Theta$, as in Setup \ref{set:GroupTransitive}. Let $\Qcal$ be an open exponential family of distributions, on which $G$ acts transitively by pushforwards.  Therefore $\Qcal$ is included in a harmonic exponential family given by a group representation $\pi$ and a $1$-cocycle $\aalpha$ as in Setup \ref{set:grpReprCocycles}. In this setting both BLR given by~\cref{eq:BLR1} and LG-BLR given by~\cref{eq:gupdate} (or~\cref{eq:theGroupUpdateInLambdas} in the natural parameters) exist.
	
	Then, the linearization of the LG-BLR update in the step size is the BLR update.
\end{theorem}

\subsection{A Final Remark on Momentum}
\label{subsec:finalmomrem}

Applying momentum, or any other optimization method that accumulates past gradients, either implicitly or explicitly requires a notion of transporting tangent vectors living in a-priori different vector spaces. We discuss in~\Cref{app:connection} the role of flat affine connections and parallel transports in this regard at a general mathematical level.

In applying momentum in the Lie group setting, we accumulate the Lie algebra vectors $\hat{Y}_t$, which is different from accumulating the natural gradients when working with exponential families. Indeed from Proposition~\ref{prop:FisherInverseGradient} we see that $\widetilde{\nabla} \mathcal{E}(q_{\llambda})\big|_{\llambda_t} = R_t \hat{Y_t}$. Even though $R_t$ is linear, it depends on the current point $g_t$ and therefore accumulating the natural gradients themselves will result in a different vector. The Lie group momentum  keeps track of and accumulates $\hat{Y}_t = R_t\inv \widetilde{\nabla} \mathcal{E}(q_\llambda) \big|_{\llambda_t}$, taking one of the preimages if $R_t$ is not invertible (In the Gaussian case these $\hat{Y}_t$ are given by the pairs $(Y_t, \vec{y}_t)$ in~\Cref{eq:gaussian_y_update,eq:gaussian_Y_update}). Then if we do the linearization as above, the update at timestep $T$ becomes  for $0< \beta< 1$
$$
\llambda_{T + 1} = \llambda_T- \alpha  R_T\left( \sum_{t = 1}^T \beta^{T- t}(1-\beta) \hat{Y}_t\right).
$$ 
This kind of momentum corresponds to using the flat connection on the Lie group that assigns left-invariant vector fields as parallel. This particular Lie group connection is nice because under its parallel transports the length and the angle between the tangent vectors with respect to the Fisher information metric stay unchanged. 

However we note that it is not the only kind of connection, and therefore not the only kind of momentum that can be applied on this exponential family. Directly accumulating the natural gradients $\widetilde{\nabla}_\llambda$ corresponds to using the $e$-connection. We also note what the mixture connection corresponds to in~\Cref{app:connection} when learning on an exponential family. In all cases further stabilizing or adaptive operations can be applied to the accumulated momenta as well---as is done with, say, AdamW or Lion optimizers.

\section{Discussion}

In this work we have shown that LG-BLR is an extension of BLR in a very precise sense. We also introduced the notion of harmonic exponential families to the field of variational inference. These characterize the families of distributions with the desirable structure of a Lie group action, which we saw gives efficient gradient estimation and updates that always yield a valid (normalizable) distribution.

The LG-BLR is extensible, and promises many new capabilities. For example in \cite{nishidalog} the authors take the LG-BLR for the group of positive reals under multiplication, the family of log-normal distributions, and extend the gradient accumulation method of Lion~\citep{chen2023symbolic}, adding regularizers relevant to the family and the group. The resulting optimization method performs stably under low precision floating point representations, and performs well compared to other multiplicative update algorithms. As another possible extension, the use of the Lie Group exponential is but one choice of a retraction, and can be replaced with easier to compute variants that can be obtained from the group exponential.

With this work, we hope that the reader has a better understanding of the LG-BLR both as a standalone method and in how it relates to the existing variational inference optimizer BLR.  

\vskip 0.2in
\bibliography{biblio.bib}

\acks{This work is supported by the Bayes duality project, JST CREST Grant Number JPMJCR2112. Eren Mehmet K{\i}ral acknowledges the support of JSPS, KAKENHI Grant Number 26K14955 titled ``\emph{Learning with Distributions and Lie Groups}''. Thomas M\"{o}llenhoff acknowledges the support of JSPS KAKENHI Grant Number 26H02541 titled ``\emph{Variational Learning for Training Better AI Models at Lower Cost}''.}

\bigskip

{\noindent \bf Author Contribution Statement}

\smallskip

\noindent Authors list: Eren Mehmet Kıral (EMK), Thomas Möllenhoff (TM), Koichi Tojo (KT),  Mohammad Emtiyaz Khan (MEK), Kenichi Bannai (KB).

\smallskip 
EMK led the project and wrote most of the manuscript; EMK and TM developed the LG-BLR update; EMK, TM and KT formulated the connection with harmonic exponential families; MEK provided insights on Bayesian Learning; and EMK, TM, KT, KB discussed the mathematical content and shaped the outline of the manuscript. MEK and KB provided the funding throughout the project. The final manuscript was approved by all authors. 
\appendix

\section{The Lie Group Update Rule}
\label{app:liegroupblr}
	
This appendix gives the mathematically careful solution to the constrained maximization problem in~\Cref{eq:FastestAscentOptimization}, with the bilinear form
\begin{equation}\label{eq:FisherBilinearOnGroup}
	B(X,Y) := \omega_{\text{Fisher}}^q (\dd (\varphi \circ L_g)\big|_e[X], \dd (\varphi\circ L_g)\big|_e[Y]) \qquad \text{ for } q = q^g.
\end{equation}
Because the Fisher metric is invariant under reparametrizations the right hand side is independent of the choice of $g$ and therefore we are justified in not writing a $g$ dependence on the left hand side. The bilinear form $B$, which is the pullback of the Fisher metric, is not necessarily nondegenerate, even though the Fisher metric is, i.e. the matrix representing $B$ may be not invertible. But regardless, we will show below that the problem in~\Cref{eq:FastestAscentOptimization} is solvable. In \cite{KiKh23} this was not shown in full generality, where we provided solutions only for the groups that were presented. 

\subsection{The Differential of the Variational Objective}
\label{app:subsec:differential}

For the sake of completeness we calculate the differential of $\mathcal{E}$ in~\cref{eq:BLP} here. The entropy's dependence on $g$ is seen to be
\begin{align*}
	-\mathcal{H}_\nu (q^{g}) &= \int_{\Theta} q^g(\ttheta) \log q^g(\ttheta) \dd\nu(\ttheta) = \int_{\Theta} \frac{1}{\chi(g)} q_0(g\inv \cdot \ttheta) \log\big(\tfrac{1}{\chi(g)} q_0(g\inv \cdot \ttheta)\big) \dd \nu(\ttheta)\\
	&= \int_{\Theta} q_0(\ttheta) \log\big(\tfrac{1}{\chi(g)} q_0(\ttheta)\big) \dd\nu(\ttheta) = - \log \chi(g)  - \mathcal{H}_\nu(q_0).
\end{align*}
Here the second line is obtained by a change of variables $\ttheta \mapsto g \cdot \ttheta$. Now we can compute that $\tfrac{\dd}{\dd t} \left(-\mathcal{H}_\nu(q^{g \exp(tX)})\right) \big|_{t = 0} = - \frac{\dd}{\dd t} \log \chi(\exp(tX)) =: \xi(X)$. 

Assume that $\Theta$ is embedded in the Euclidean space so that we can take its gradient $\nabla \ell$ with respect to $\ttheta$. The directional derivative of the expectation term can also be computed with a change of variables. First let us compute the expectation of the loss under the perturbed distribution
\begin{align*}
\EE_{q^{g\exp(tX)}}[\ell] &= \int_{\Theta} \ell(\ttheta) q^{g \exp(tX)}(\ttheta) \dd\nu(\ttheta) \\
&= \int_{\Theta}  \ell(\ttheta)  \frac{1}{\chi(g \exp(tX))} q_0((g\exp(tX))\inv \cdot \ttheta) \dd\nu(\ttheta)\\
&= \int_{\Theta}  \ell(\ttheta)  \frac{1}{\chi(\exp(tX))} \frac{1}{\chi(g)} q_0(g\inv ((g\exp(tX)g\inv)\inv \cdot \ttheta)) \dd\nu(\ttheta)\\
&= \int_{\Theta} \ell(\ttheta) \frac{1}{\chi(\exp(tX))}q^g((g \exp(tX) g\inv)\inv \cdot \ttheta) \dd \nu(\ttheta)\\
 &= \int_{\Theta} \ell\left((g \exp(tX) g\inv) \cdot \ttheta \right) q^g (\ttheta) \dd\nu(\ttheta).
\end{align*}
Where in the last line we make the change of variables $\ttheta \mapsto (g \exp(tX) g\inv )  \cdot \ttheta$. The multiplier $\chi$ exactly accounts for the Jacobian determinant in this change of variables formula. Notice that now the only $t$ dependence is in the $\ell$ function and if we take the derivative at $t = 0$,
\[
	\int_{\Theta} \nabla \ell(\ttheta)^\top (\operatorname{Ad}_{g}(X) \cdot \ttheta) q^g(\ttheta) \dd\nu(\theta).
\]
This is an expectation that can be approximated with stochastic sampling from $q^g$, which in turn is equivalent to sampling $\ttheta_0 \sim q_0$ and applying $g \cdot \ttheta_0$. 

Here $\operatorname{Ad}_g(X)$ is an element of the Lie algebra given by a differential, and can be calculated exactly by $\Ad_g(X) := \frac{\dd}{\dd t}(g\exp(tX) g\inv)\big|_{t = 0}$, and $\Ad_g(X) \cdot \ttheta$ refers to the infinitesimal action of this Lie algebra vector on $\ttheta$, defined as a limit, i.e. for $Z \in \mathfrak{g}$ the infinitesimal action is given as $Z \cdot \ttheta := \tfrac{\dd}{\dd t} (\exp(tZ)\cdot \ttheta)\big|_{t = 0} \in T_{\ttheta} \Theta$. The gradient $\nabla \ell(\theta)$ is actually coming from the cotangent vector in $T_\ttheta^*\Theta$ so these two quantities can be paired. But as we assumed without loss of generality that $\Theta$ is embedded in $\R^n$, the pairing is simply the dot product between two vectors in $\R^n$.

\subsection{Solving for the Gradient} \label{app:subsec:gradient}

Let us show that we can always compute the gradient of $\mathcal{E}(q^g)$ with respect to the positive semidefinite metric. With a slight abuse of notation, let $B, P = P_g$ and $F$ denote matrices corresponding to the bilinear form in~\Cref{eq:FisherBilinearOnGroup} on $\mathfrak{g}$, the map $\dd (\varphi \circ L_g)\big|_e$ and the Fisher metric $\omega_{\text{Fisher}}^{q^g}$ after a choice of bases in $\mathfrak{g}$ and $T_{q^g}\Qcal$. Note that if~\Cref{eq:FisherBilinearOnGroup} is written in coordinates, corresponds to the equality $B = P^\top F P$.

Writing the optimization following~\Cref{eq:FastestAscentOptimization} in coordinate form, the constrained optimization problem becomes
\[
	\dd(\mathcal{E} \circ \varphi \circ L_g)\Big|_e [\mathbf{X}] = \zzeta^\top  P\mathbf{X} \qquad \text{ subject to } \qquad \mathbf{X}^T B \mathbf{X} \leq 1.
\]
where $\zzeta$ is the column-vector form of the linear functional $\dd\mathcal{E}\big|_{q^g} : T_{q^g} \Qcal \to \R$ consisting of partial derivatives in the given basis, and $\mathbf{X}$ is a column vector representation of a Lie algebra element in $\mathfrak{g}$. 

Using the method of Lagrange multipliers, we solve  
\[
	\argmax_{\mathbf{X} \in \mathfrak{g}} \quad  \zzeta^\top P \mathbf{X} + \beta  \mathbf{X}^\top B \mathbf{X}.
\]
The first derivative in $\mathbf{X}$ yields the equality $\zzeta^\top P +  2\beta \mathbf{X}^\top B = 0$ or in other words
\[
	2\beta P^\top F P\mathbf{X} = - P^\top \zzeta
\] 
which can be solved for all $\mathbf{X}$ satisfying $P \mathbf{X} = -\frac{1}{2\beta} F\inv \zzeta$. Since $P$ is the matrix of the differential of $\varphi$ which parametrizes $\Qcal$, it is necessarily surjective, so we can surely solve for $\mathbf{X}$, though not necessarily uniquely. This equation may have more than one solution if $\dim \Qcal < \dim \mathfrak{g}$, and in that case we just pick any one of the solutions. 

Even though in showing that there exists at least one solution to~\Cref{eq:FastestAscentOptimization} we used the invertibility of $F$, one does not need to invert the Fisher matrix to do this calculation. One can simply solve the system of equations $B\mathbf{X} = -\frac{1}{2\beta} P^\top \zzeta$, and $P^\top \zzeta$ is the vector corresponding to $\dd(\mathcal{E} \circ \varphi \circ L_g)$ after a choice of basis of $\mathfrak{g}$, and $B$ needs to be calculated in a basis of $\mathfrak{g}$ once and for all, it does not change throughout the optimization.

Furthermore, using the calculations in~\Cref{app:subsec:differential} we can directly calculate the components of $ P^\top \zzeta$ (without calculating those of $\zzeta$) via the formula  
\begin{equation*}
\frac{\dd}{\dd t} \mathcal{E}(q^{ge^{t\mathbf{X}}})\Big|_{t = 0}=  \EE_{q^g \dd \nu} [\left(\nabla_\ttheta \ell (\ttheta)\right)^\top  (\operatorname{Ad}_g(\mathbf{X}) \cdot \ttheta)] + \tau \xi(\mathbf{X})
\end{equation*}
by substituting $\mathbf{X} = \mathbf{X}_i$, given a basis $\{\mathbf{X}_i\}_i$ of the Lie algebra.

\subsection{Linearization of LG-BLR for Curved Exponential Families}
\label{app:subsec:curvedlinearization}

For this subsection assume that the family $\Qcal$ is not necessarily open. It is a curved exponential family closed under the Lie group action. But the orbit of $\llambda_0$ under $\pi^*$ doesn't form an open set in $\Omega \subseteq \dualsp$ as depicted in~\Cref{fig:Gorbit}. 

\begin{figure}[h!]
\centering
\includegraphics[width = 0.35\textwidth]{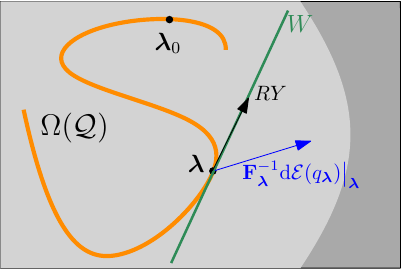}
\caption{The Lie group BLR can only move along $\Qcal$, but inside the general exponential family $\EF(\alpha)$ the natural parameters can be updated in directions that have more degrees of freedom.}
\label{fig:ProjectionDirection}
\end{figure}

 Here when we linearize the LG-BLR update, it is clear that the range of motion afforded to us by the Lie group Bayesian learning rule is restricted. Rather than being able to move in any which direction in $\Omega \subseteq \dualsp$, we can only move along $W = T_\llambda \Omega(\Qcal)$, see~\Cref{fig:ProjectionDirection}. 

%
%We write the direction only version of the "fastest ascent" problem. \begin{equation}\label{eq:FastestRestrictedDirection}
%\hat{Y} = 	\argmax_{\substack{\mathbf{v} \text{ such that} \\ \mathbf{v} = RX,  X\in \mathfrak{g}}} \dd (\mathcal{E} \circ \iota )[\mathbf{v}]  \qquad \text{ subject to } \quad \omega_{\text{Fisher}}^\llambda(\mathbf{v},\mathbf{v}) = 1.
%\end{equation}
%

\begin{proposition}\label{prop:FisherInverseGradientProjected}
	Assume the setting and notation of~\Cref{subsec:dirMotion}, let $\Qcal$ be an exponential family where $\Omega(\Qcal)$ is an (immersed) submanifold of $\Omega$. Let $W:= T_\llambda (\Omega(\Qcal)) \subseteq \dualsp$, and let $\hat{Y} \in \mathfrak{g}$ be the update direction, satisfying $B(\hat{Y}, X) = \dd(\mathcal{E} \circ \varphi \circ L_g)\big|_{e}[X]$ for all $X \in \mathfrak{g}$. Then,
	\[
		 R\hat{Y} = \Proj_{W} \widetilde{\nabla}_\llambda \mathcal{E}(q_\llambda)
	\]
	where the orthogonal projection is taken with respect to the inner product given by the Fisher information metric $F_\llambda$. In particular if $\Qcal$ is an open exponential family, we have $T_\llambda \Omega(\Qcal) = \dualsp$ in which case $R\hat{Y} = \widetilde{\nabla}_\llambda \mathcal{E}(q_\llambda) = F_\llambda\inv \dd (\mathcal{E} \circ \iota) \big|_{\llambda}$.  
\end{proposition}

\begin{proof}
First note that since $\mathbf{N}$ traces out the natural parameters in $\Omega(\Qcal)$, we have that the image of $R = \dd( \mathbf{N} \circ L_g)\big|_{e}$ is $W$. From the definition of $B$ as a pullback, we get
\[
	\dd(\mathcal{E} \circ \iota)\big|_{\llambda} [RX] = \dd (\mathcal{E}\circ \varphi \circ L_g)\big|_{e}[X] = B(\hat{Y}, X) = \omega_{\text{Fisher}}^\llambda (R\hat{Y} , R X) 
\]
So for all $\vw \in W$, the Lie algebra vector $\hat{Y}$ satisfies $ \omega_{\text{Fisher}}^\llambda (R\hat{Y}, \vw) = \dd(\mathcal{E} \circ \iota)\big|_{\llambda}[\vw] $. Furthermore by the definition of natural gradient, for all $\vv \in \dualsp$ the equality
\[
	\omega_{\text{Fisher}}^\llambda (\widetilde{\nabla}_\llambda \mathcal{E}(q_{\llambda}), \vv) = \dd(\mathcal{E} \circ \iota)\big|_{\llambda}[\vv]
\]
holds. This means that $\omega_{\text{Fisher}}^\llambda (R\hat{Y} - \widetilde{\nabla}_{\llambda} \mathcal{E} (q_\llambda), \vw) = 0$ for all $\vw\in W$, meaning that the difference is in $W^\perp$, proving the proposition. 
\end{proof}

\section{Momentum and Connections}
\label{app:connection}

When using gradient methods on a differentiable manifold $M$ (including the standard parameter space $M = \Theta = \R^P$) the act of applying momentum, or other acceleration techniques for optimization, involves accumulating past gradients. These gradients are on different tangent spaces; $\vec{g}_{\text{past}} \in T_{\vec{p}_{\text{past}}} M$ vs.\ $\vec{g}_{\text{now}} \in T_{\vec{p}_{\text{now}}} M$. In order for this accumulation to make sense, we require a way to connect different tangent spaces, i.e. an affine connection.

An affine connection is defined as an operator taking in two vector fields $X,Y\in \mathfrak{X}(M)$ and giving another vector field $Z = \nabla_X Y$ satisfying certain conditions. $Z_{\vec{p}}$---the resulting vector field's value at a point $\vec{p}$---is a measure of the change of $Y$ along $X$, infinitesimally at $\vec{p}$. For our purposes of connecting different tangent spaces, however, the equivalent notion of parallel transport is more relevant.

Given an affine connection $\nabla$ and a smooth path $\gamma : [0, 1] \to M$, the parallel transport along $\gamma$ is a linear map
$$P_{\gamma } : T_{\gamma(0)} M \to T_{\gamma(1)}M$$
that carries the tangent vectors by solving a differential equation, $\nabla_{\dot{\gamma}(t)} V(t) = \vec{0}$ to be precise. The interpretation of the differential equation is that according to the affine connection $\nabla$, the vectors $V(t)$ stay constant along the path $\gamma$. Setting $V(0) = \vec{v} \in T_{\gamma(0)}M$ as the initial condition, we get a path of tangent vectors $V(t)$ transporting $\vec{v}$ to $V(1) \in T_{\gamma(1)}M$, which we define to be the parallel transport $P_\gamma(\vec{v})$. 

In general, parallel transports depend on the path $\gamma$ along which vectors are transported. However if the curvature tensor\footnote{Defined as $R(X,Y)Z = (\nabla_X \nabla_Y - \nabla_Y\nabla_X - \nabla_{[X,Y]})Z$ for three vector fields $X,Y,Z$ and where $[X,Y]$ is the Lie bracket of the two vector fields.} for the given connection is identically zero---in which case we call this a \emph{flat connection}---then for two paths with the same endpoints that can be deformed into each other their parallel transports are the same. This means that, unless there is a global topological obstruction such as in the handle of a coffee mug, for a flat connection we can talk about \emph{the parallel transport}. See, e.g.\ p.36 in~\citet{Heckman2019}.

For a Lie group $G$, we can declare left invariant vector fields to be parallel by setting $\nabla_X Y = \vec{0}$ for $X,Y$ left invariant\footnote{A left invariant vector field means that the differential of the left-multiplication-by-$g$ map at $p$, i.e. the linear map $\dd L_g\big|_p :T_pG \to T_{gp}G$, sends the vector $X_p \in T_pG$ to $X_{gp}\in T_{gp}G$.}. This is enough to define a flat affine connection. Flatness follows directly from the definition of the curvature tensor using that the Lie bracket of two left-invariant vector fields is also left-invariant. The parallel transports along all paths (locally in contractible patches) are given by the maps $\dd L_g$. 

In the LG-BLR we expressed our updates at any point $g_t$ by $g_{t + 1} = g_t \exp(-\eta\hat{Y}_t)$ for an element $\hat{Y}_t \in \mathfrak{g}$. Given the flat connection defined in the above paragraph, the identification of the gradient at $g_t$ to the vector $\hat{Y}_t$ in the Lie algebra $\mathfrak{g} = T_eG$ corresponds to the parallel transport. As a side note, we remark that the retraction used in this update given by the exponential map is a geodesic for this connection.

The exponential families $\EF_{\nu,\text{\vT}}$ famously have two flat connections $\nabla^e$ and $\nabla^m$\footnote{Further, these two connections are dual to one another under the Fisher information metric, hence the name dually-flat connections, but we will not use this part of the structure.}. The parallel transport with respect to the $\nabla^e$ connection corresponds to the flat vector space connection in the $\llambda$ space. In other words for two natural parameters $\llambda_1, \llambda_2 \in \Omega_{\nu, \text{\vT}} \subseteq \dualsp$ the parallel transport given by the $e$-connection allows us to identify the spaces where the natural gradients $\widetilde{\nabla}_{\llambda} \mathcal{E}(\llambda_1)$ and $\widetilde{\nabla}_{\llambda} \mathcal{E}(\llambda_2)$ live, which are technically in different tangent spaces $T_{\llambda_1} \Omega$ and $T_{\llambda_2}\Omega$. The concrete coordinates already realize them as vectors in $\dualsp$, which coincides with the identification of the tangent spaces using the $\nabla^e$ parallel transport. The parallel transport given by the $m$-connection, on the other hand, identifies space of the gradients in the expectation parameters $\vec{\mu}$, which coincides with the identification given by the concrete coordinatization $\dd \mathcal{E}\big|_{\llambda_1}, \dd \mathcal{E}\big|_{\llambda_2}\in \primsp$. These two are related to each other by the Fisher metric $F_\llambda$.

Implementing the momentum as the Exponential Moving Average (EMA) of past gradients (parallel transported to the current point's tangent space with respect to a chosen affine connection) we thus have three notions of such an EMA: 
\begin{itemize}
\item For the $e$-connection the EMA is simply accumulating the natural gradients: $$\vec{m}_t = \beta \vec{m}_{t -1} +(1-\beta) \widetilde{\nabla}_\llambda \mathcal{E}(\llambda_t).$$ This accumulation is happening in the natural parameter space $\dualsp$.
\item For the $m$-connection the parallel transport means that we apply an EMA on the ordinary gradients (not natural gradient), and the momentum in the natural parameter space is obtained using the Fisher information metric at the final (current) point:
$$
	\vec{n}_t = \beta \vec{n}_{t -1} +  (1 - \beta) \dd \mathcal{E}\big|_{\llambda_{t}}
\qquad \vec{m}_t = F_{\llambda_{t}}\inv \vec{n}_t $$
\item For a harmonic exponential family, the statistical manifold can be parametrized by the Lie group and we can apply the left-invariant connection described above. This corresponds to an EMA of the update vectors  $\hat{Y}_t \in \mathfrak{g}$. For an exponential family, the relation of such an averaging to calculations in the $\llambda$ coordinates is described in more detail in  \cref{subsec:finalmomrem} of the main text.
\end{itemize}

\section{Background on Exponential Families}
\label{app:ExponentialFamily}

\subsection{Non-constructive Definition of an Exponential Family}
Let $\mathcal{R}(X)$ denote the set of all Radon measures on a manifold  $X$.
\begin{definition}[Exponential Family]\label{def:exponential_family}
  A non-empty subset $\mathcal{P}\subset\mathcal{R}(X)$ consisting of probability measures is an \emph{exponential family on $X$} if 
there exists a pair $(\nu,\vT)$ satisfying 
\begin{enumerate}[label=$(\roman*)$]
\item $\nu\in \mathcal{R}(X)$, 
\item $\vT \colon X \to \primsp$ is a continuous map, 
\item where $\primsp$ is a finite dimensional real vector space, 
\item and for any probability distribution $p \in \mathcal{P}$, there exists $\llambda \in \dualsp$ such that the Radon-Nikodym derivative of $p$ with respect to $\nu$ is given by
\begin{align}
\frac{\dd p}{\dd\nu}(\vx)&=\exp(-\pairing{\llambda}{\vT(\vx)}-A(\llambda)), \label{eq:exponential} \\
\text{where } A(\llambda)&:=\log\int_{X} \exp(-\pairing{\llambda}{\vT(\vx)})d\nu(\vx) \quad (\text{log-partition function}). 
\end{align}
\end{enumerate}
Here, $\dualsp$ denotes the dual space of $\primsp$ and $\pairing{\cdot}{\cdot}\colon \dualsp\times \primsp\to \R$ denotes the natural pairing. 
We call the pair $(\nu, \vT)$ a \emph{realization} of $\mathcal{P}$. 
Moreover, if the dimension of $\primsp$ is minimal among all the realizations of $\mathcal{P}$, the realization is said to be \emph{minimal}. 
\end{definition}

\begin{remark}
  The definition above mainly relies on {\cite[\S~8.1]{bn78}}. 
In {\cite[\S~8.1]{bn78}} the notion of an exponential family is defined in the category of measurable spaces and measurable maps. 
On the other hand, we consider exponential families in the category of topological spaces and continuous maps in order to prove Lemma~\ref{fact:Tojo-Yoshino_GSI25-paper}. 
In our setting, we do not need to assume that $\nu$ is $\sigma$-finite (see the appendix in \cite{ToYo21}). 
\end{remark}

Given an exponential family $\mathcal{P}$, and a realization $R = (\nu, \vT)$ of $\mathcal{P}$ denote by $\Omega_R(\mathcal{P}) \subseteq \Omega_R \subseteq \dualsp$ the set of natural parameters $\llambda$ that give elements of $\mathcal{P}$. It is clearly possible that more than one $\llambda$ yields the same distribution if the components $\{1, T_1(\ttheta), \ldots, T_d(\ttheta)\}$ are not linearly independent functions (up to $\nu$-almost everywhere equivalence). In such a case, clearly $R$ is not a minimal realization of $\mathcal{P}$ since we can replace $\vT$ by $\vT'$ by omitting certain dimensions so that the coordinate functions are linearly independent. Another source of non-minimality can occur if $\Omega_R(\mathcal{P})$ lies in an affine space. Note first that the affine hull of these natural parameters of $\mathcal{P}$ is then of the form $\llambda_b + \mathrm{W}$ for some subspace $\mathrm{W} \subseteq \dualsp$. It is possible to change the realization $R' = (\nu e^{-\langle \llambda_b, \vT \rangle}, \vT')$ where $\vT'$ is a subset of the linear combination of functions of $\vT$. In this case we get a minimal representation of the exponential family. 

The log-partition function $A$ is differentiable in the interior of $\Omega_{R}$, and for $\llambda \in \Omega_R^\circ \subseteq \Omega_R$ in the interior, the negative gradient of the log partition function is the change of coordinates function from the natural parameters to the mean parameters
\begin{align*}
	-\nabla A : \Omega^\circ &\longrightarrow \primsp \\
					\llambda &\longmapsto \mmu := \EE_{q_{\llambda}}[\vT].
\end{align*}
Here the mean variable $\mmu = \EE_{q_\llambda}[\vT] \in \primsp$ is another coordinatization of such distributions with $\llambda \in \Omega_R^\circ$. 
For a minimal realization this map from the interior is one-to-one, it is a change of coordinates.

\subsection{The Gradient vs the Differential from a Geometric Point of View} \label{app:subsecGradvsDiff}

There is a meaningful distinction between the gradient and the differential of a function at a point. Let us take, for concreteness, the example of $\mathcal{E}: \Omega(\mathcal{P}) \to \R$. The differential of $\mathcal{E}$ at a $\llambda$ is actually a linear map
\[
	\dd \mathcal{E}\big|_{\llambda} : T_\llambda \Omega(\mathcal{P}) \cong \dualsp \to T_{\mathcal{E}(\llambda)} \R \cong \R.
\]
In other words $\dd \mathcal{E}\big|_{\llambda} \in (\dualsp)^* \cong \primsp$. This is not the same space where $\llambda$ lives, it is its dual. In other words if $\llambda$ are represented by column vectors, then the differential $\dd \mathcal{E}\big|_\llambda$ is a row vector, representing a linear functional on the $\llambda$'s. It is a cotangent vector, and thus cannot be treated as a tangent direction to perform gradient descent on the manifold directly. 

A trivial way to convert the differential to a tangent vector in coordinates is to take the transpose of the row vector, turning it into a column vector. This corresponds to the special case of choosing a Euclidean metric in the following more general construction. A metric is a bilinear form on the tangent space. Again for concreteness consider the Fisher metric at a point $\llambda \in \Omega(\mathcal{P})$. This is a bilinear form $F_\llambda : \dualsp \times \dualsp \to \R$ which can also be seen as a linear function $F_\llambda : \dualsp \to \primsp$ upon fixing one of the variables (also called currying in computer science). If the metric is nondegenerate, i.e., if there are no nonzero vectors which are orthogonal to every other vector (including itself), then this map is invertible and $F_\llambda\inv$ gives us a method to convert the differentials (which are cotangent vectors at $\llambda$) to tangent vectors. In fact
\[
	\argmax_{\vv \in T_\llambda \Omega(\mathcal{P}) \cong \dualsp} \dd \mathcal{E}\big|_\llambda [\vv] \qquad \text{ subject to } \qquad \|\vv\|_{\text{Fisher}}^2 = \omega^{\text{Fisher}}(\vv,\vv) = \vv^\top F_\llambda \vv= 1.
\]
is solved by $\vv/\|\vv\|_{\text{Fisher}}$ where $\vv = F_\llambda\inv \dd \mathcal{E}\big|_\llambda$. 

The Fisher metric at $\llambda \in \Omega(\mathcal{P})$ is a bilinear form on the tangent space $T_\llambda \Omega(\mathcal{P}) \cong \dualsp$ an analogue of the dot product on the tangent vectors. Concretely it is defined as 
\begin{align*}
	\vF_{\text{$\vlambda$}} &: \dualsp \times \dualsp \longrightarrow \bbR\\
	\vF_{\text{$\vlambda$}}(\vv, \vw) &= \EE_{\vparam \sim q_\llambda}[(\partial_{\mathbf{v}}\log q_\llambda)(\partial_{\mathbf{w}}{\log q_\llambda})]. 
\end{align*}

This is similar to what happens in the Lie algebra. With $B$ on $\mathfrak{g}$ we can rewrite~\Cref{eq:FastestAscentOptimization} as finding a $\hat{Y} \in \mathfrak{g}$ that satisfies $B(\hat{Y}, \cdot) = \dd (\mathcal{E}\circ \varphi \circ L_g)\big|_e$. Here both sides are elements of $\mathfrak{g}^*$, covectors taking in $X \in \mathfrak{g}$ and spitting out numbers. Such a $\hat{Y}$ can be found if the linear functional on the right hand side vanishes on the kernel of $B$. The equivalent presentation involving an optimization constrained to the unit ball also builds the intuition explaining the phrase ``the direction of fastest ascent/descent''.

\end{document}